\documentclass[acmtog]{acmart}
\acmSubmissionID{310s1}

\usepackage{booktabs} 

\usepackage[ruled]{algorithm2e} 

\SetAlFnt{\small}
\SetAlCapFnt{\small}
\SetAlCapNameFnt{\small}
\SetAlCapHSkip{0pt}

\usepackage{bm}

\newcommand{\ww}[1]{#1} 
\newcommand{\WW}[1]{}
\newcommand{\GR}[1]{}
\newcommand{\gr}[1]{}
\newcommand{\CO}[1]{}
\newcommand{\hx}[1]{}
\newcommand{\FZ}[1]{}

\newcommand{\ZL}[1]{}

\newcommand{\spl}[1]{#1}

\usepackage{alphabeta}
\usepackage{pifont}
\usepackage{subcaption}
\usepackage{graphicx}
\usepackage{amsmath}
\usepackage{bm}
\usepackage{nicematrix}

\DeclareMathOperator{\MLP}{MLP}

\usepackage{booktabs}
\usepackage{gensymb}
\usepackage{colortbl}
\usepackage{soul}
\usepackage{setspace}
\usepackage{float}

\definecolor{first}{rgb}{1,0.85, 0.7}
\definecolor{second}{rgb}{1,1, 0.8}
\definecolor{firstalt}{rgb}{1, 0.7, 0.7}

\newcommand{\first}{{\cellcolor{first}}}

\newcommand{\second}{{\cellcolor{second}}}

\acmJournal{TOG}

\begin{document}
\title{Gaussian Head \& Shoulders: High Fidelity Neural Upper Body Avatars with Anchor Gaussian Guided Texture Warping}

\author{Tianhao Wu}
\affiliation{%
 \institution{University of Cambridge}
 \country{United Kingdom}
}

\author{Jing Yang}
\affiliation{%
 \institution{University of Cambridge}
 \country{United Kingdom}
}

\author{Zhilin Guo}
\affiliation{%
 \institution{University of Cambridge}
 \country{United Kingdom}
}

\author{Jingyi Wan}
\affiliation{
    \institution{University of Cambridge}
    \country{United Kingdom}
}

\author{Fangcheng Zhong}
\affiliation{%
 \institution{University of Cambridge}
 \country{United Kingdom}
}

\author{Cengiz Oztireli}
\affiliation{%
 \institution{Google Research,}
 \institution{University of Cambridge}
 \country{United Kingdom}
}



\begin{abstract}
The ability to reconstruct realistic and controllable upper body avatars from casual monocular videos is critical for various applications in communication and entertainment. By equipping the most recent 3D Gaussian Splatting representation with head 3D morphable models (3DMM), existing methods manage to create head avatars with high fidelity. However, most existing methods only reconstruct a head without the body, 
substantially limiting their application scenarios.
We found that naively applying Gaussians to model the clothed chest and shoulders tends to result in blurry reconstruction and noisy floaters under novel poses. This is because of the fundamental limitation of Gaussians and point clouds -- each Gaussian or point can only have a single directional radiance without spatial variance, 
therefore an unnecessarily large number of them is required to represent complicated spatially varying texture, even for simple geometry.
In contrast, we propose to model the body part with a neural texture that consists of coarse and pose-dependent fine colors. To properly render the body texture for each view and pose without accurate geometry nor UV mapping, we optimize another sparse set of Gaussians as anchors that constrain the neural warping field that maps image plane coordinates to the texture space. We demonstrate that Gaussian Head \& Shoulders can fit the high-frequency details on the clothed upper body with high fidelity and potentially improve the accuracy and fidelity of the head region. We evaluate our method with casual phone-captured and internet videos and show our method archives superior reconstruction quality and robustness in both self and cross reenactment tasks. To fully utilize the efficient rendering speed of Gaussian splatting, we additionally propose an accelerated inference method of our trained model without Multi-Layer Perceptron (MLP) queries and reach a stable rendering speed of around 130 FPS for any subjects. 

\end{abstract}

%
%


%
%

\begin{teaserfigure}
  \includegraphics[width=\textwidth]{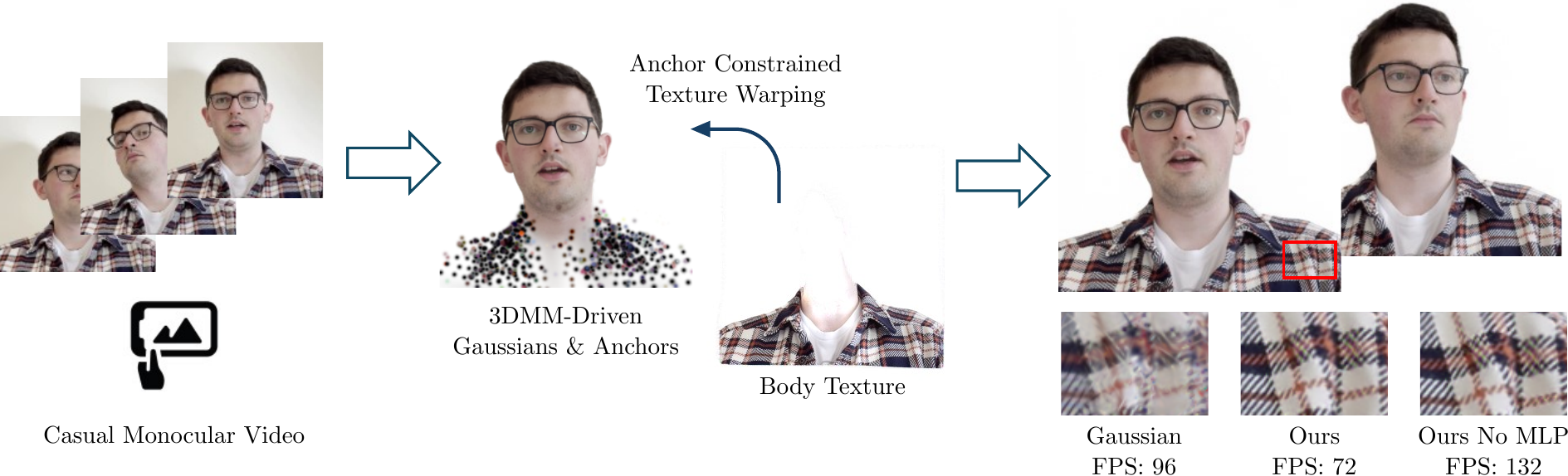}
  \caption{\textbf{Gaussian Head \& Shoulders} reconstructs 3DMM-driven upper body avatars from casual monocular videos. By utilizing a high-frequency body neural texture which is warped using a neural texture warping field constrained by a set of sparse anchor Gaussians, we can learn sharp details of the cloth texture with highly efficient rendering speed.}
  \label{fig:teaser}
\end{teaserfigure}

\maketitle


\section{Introduction}


Personalized and controllable 3D head avatar is a crucial asset for interactive Mixed Reality and metaverse applications. 
%
%
Recent developments in the 3D representations such as 3DMM \cite{FLAME:SiggraphAsia2017, bfm}, Neural Radiance Field \cite{nerf}, Instant Neural Primitives \cite{instantNGP}, and other implicit representations~\cite{occupancy_network} have brought
rapid advancements in the reconstruction of vivid and controllable neural avatars \cite{zheng2022imavatar, neuralheadavatar, INSTA, Gao2022nerfblendshape}.
With the most recent 3D Gaussian Splatting representation \cite{gaussian_splatting}, neural avatars can be convincingly
reconstructed from a monocular video with impressive fidelity.
However, most current methods for creating head avatars concentrate solely on the face and head, discarding other visible parts of the body by using a semantic mask during the training process.
Consequently, this results in avatars that appear as heads without bodies, which is not sufficient
for many immersive applications, including video conferencing, where a more complete avatar is needed~\cite{SplattingAvatar:CVPR2024, xiang2024flashavatar, INSTA,Gao2022nerfblendshape}. 
Recent techniques aim to create more complete avatars by including visible parts of the body, like shoulders and chest \cite{pointavatar, psavatar, gaussianblendshape, zheng2022imavatar}. However, they are limited to simplified settings where the subject dresses in plain clothing without detailed textures and is instructed to restrict upper body movement. On the other hand, existing full-body avatar methods typically focus on the overall quality of the limbs and torso and fail to faithfully capture the fine details such as high-frequency texture on clothes~\cite{hugs, hu2024gaussianavatar_human, li2024animatablegaussians_human, lei2023gart}. 
Applications of neural avatars that require detailed reconstruction of the upper body area often encounter significant challenges in capturing faithful and intricate details. Overall, current methods still fall short of delivering the level of performance needed for practical, real-world use.

%
The Gaussian Splatting representation, while being efficient and effective in certain aspects, struggles with accurate modeling of clothed upper bodies. As one of its fundamental limitations, each Gaussian can represent only one color from a specific viewing angle. This heavily limits its capability to handle dynamic objects that have complex textures, such as clothing with intricate patterns. To capture the detailed appearance of such objects, an excessively large number of Gaussians would be needed, increasing memory requirement and slowing down the rendering speed. 
%
In addition, the complicated pose-dependent appearances such as brightness changes and cloth wrinkles further increase the difficulty of modeling them with plain Gaussians alone.
As a result, when the reconstructed avatar is driven to novel poses, the Gaussians tend to produce several undesirable artifacts such as blurred texture, incorrect colors or floating ellipsoid; see Fig~\ref{fig:teaser}.

To address the limitations of existing Gaussian-based avatar methods on clothed upper-body region, we argue that the chest and shoulders are expected to have relatively simpler geometry and more intricate deformation compared to the head.
Therefore, modeling them with regular and 3DMM-driven Gaussians would be unsuitable and is an over-complication of the problem.
Instead, a more appropriate and standard 
approach would be representing their appearance with a high-frequency texture.

In a traditional texture-based rendering pipeline, the texture is first mapped to mesh geometry in the 3D world space via UV mapping, and then rasterized to the 2D image plane in the view space to obtain the pixel color. However, this approach requires a well-defined UV mapping and accurate mesh geometry, which is challenging to obtain from monocular videos alone due to the lack of multi-view correspondences. 
Besides, compared to faces that share more common characteristics and stronger priors, the appearance of upper bodies can vary dramatically depending on the texture and tightness of the clothes and they hence contain
fewer detectable landmarks.
Consequently, body 3DMMs such as SMPL \cite{SMPL:2015} fail to provide geometry accurate enough for this purpose.
%


Hence, we propose to bypass the mapping from texture space to world space, and instead use a sparse set of Gaussians as ``anchors" to define a direct neural warping field between the texture space and the image plane. As the tracking of body 3DMM tends to be inaccurate due to the lack of landmarks, we only transform anchor Gaussians together with the head Gaussians via a head FLAME 3DMM~\cite{FLAME:SiggraphAsia2017} through Linear Blend Skinning (LBS). The transformed anchor Gaussians are used as soft constraints of the texture warping represented by a coordinate-based MLP, which is optimized together with the neural texture, regular Gaussians, and the anchor Gaussians. As the resolution of the neural texture is not limited by the number of Gaussians or the density control scheme, we can easily learn the high-frequency textures with sharp details on the clothes and avoid the common artifacts exhibited in Gaussian rendering under novel poses; see Fig~\ref{fig:teaser}.

To maintain a competitive rendering speed with Gaussian Splatting and 
enable real-time interactive applications, we additionally propose a method to remove the neural warping field and neural texture in the model and allow inference of reconstructed avatars at novel poses without any MLP queries. This accelerated inference effectively increases the rendering speed from 70 FPS to around 130 FPS, which surpasses the rendering speed of plain Gaussian Splatting avatars for subjects with high-frequency clothes.

We evaluate the proposed method with various casual monocular videos collected using smartphones or from the Internet.
Compared to state-of-the-art methods which incorporate different representations including neural radiance field, Gaussian Splatting, and point clouds, we show that our approach achieves better performance and robustness for both self-reenactment and cross-reenactment tasks.
In summary, our contributions are:

\begin{itemize}
  \item We propose a novel approach that maps intricate texture to the image plane via a sparse set of anchor Gaussians driven by LBS with 3DMM. This allows accurate and robust modeling of high-fidelity clothed chest and shoulders with less number of Gaussians.

  \item We propose a method to remove the MLP in our method at inference time to prevent any costly queries when rendering with novel poses and expressions and reach a rendering speed of around 130 FPS.

\end{itemize}

\section{Related Works}

\paragraph{Neural Head Avatars} 
The recent advancement in neural 3D implicit and explicit representations has sparked a surge of methodologies within the field of controllable 3D head avatars. Among these approaches, a prominent family of methods involves the reconstruction of a 5D neural radiance field, manifested through various forms such as pure MLP~\cite{dynerf_for_face, composition_dynerf_for_head, kirschstein2023nersemble}, hash grid latents~\cite{xu2023avatarmav,Gao2022nerfblendshape, INSTA, xu2023avatarmav, dhamo2023headgas, xiang2024flashavatar, saito2024rgca, chen2023monogaussianavatar}, and 3D Gaussians~\cite{gaussianblendshape, psavatar}. Another set of methods utilizes more explicit representations such as deformable meshes with neural textures ~\cite{neuralheadavatar, zheng2022imavatar, varitex, icml2020_2086, Khakhulin2022ROME, kim2018deep_video_portrait} and point clouds~\cite{pointavatar}. 
Notably, the point cloud representation used by PointAvatar~\cite{pointavatar} is similar
to 3D Gaussian Splatting in 
their modeling of point size, individual point colors, and opacities that decay with the distance to point center. 
The core differences between the two are that Gaussian Splatting additionally supports anisotropic Gaussians, individual Gaussian size, view-dependent appearance via SH coefficients, and a more sophisticated density control scheme that grows the number of Gaussians. 
%
%
%
With the most recent Gaussian Splatting techniques, the head avatars reconstructed from monocular videos already reach high fidelities. 
However, many methods simplify the problem by reconstructing only the head and neck part, resulting in a head-only reconstruction that is not suitable for many applications. 
Several methods have attempted to also model the chest and shoulders to provide a more immersive user experience~\cite{zheng2022imavatar, psavatar, gaussianblendshape, pointavatar}. However, they are limited to simple clothes with plain colors, and cannot handle the movements in the upper body in the video. 

\paragraph{Neural Full-Body Avatars}
Several works have tried to reconstruct a controllable full-body neural avatar from multi-view or monocular videos~\cite{liu24-GVA, SplattingAvatar:CVPR2024, svitov2024haha, li2024animatablegaussians_human, lei2023gart, hugs, hu2024gaussianavatar_human, jiang2022neuman}.
Due to the highly articulated nature of human bodies, they tightly rely on body 3DMMs to deform the neural body representation via LBS. 
However, they typically fail to faithfully capture subjects with complicated or loose clothing as those cannot be modeled with existing body 3DMMs.
Li et. al.~\cite{li2024animatablegaussians_human} first reconstruct an SDF field as parametric models, then incorporate a StyleUnet to directly infer parameters of 3D Gaussians in the canonical space and deform it to each pose via LBS. They can reconstruct high-quality avatars even under cases of loose clothing, but require synchronized multiview videos as input. 
As methods that reconstruct animatable full-body avatars typically focus on the overall quality of the torso and limbs, they tend to present non-trivial artifacts when reconstructing and re-animating an avatar that has a tight focus around the head and shoulder regions.
%

\section{Method}

\begin{figure*}
    \centering
    \includegraphics[width=0.95\textwidth]{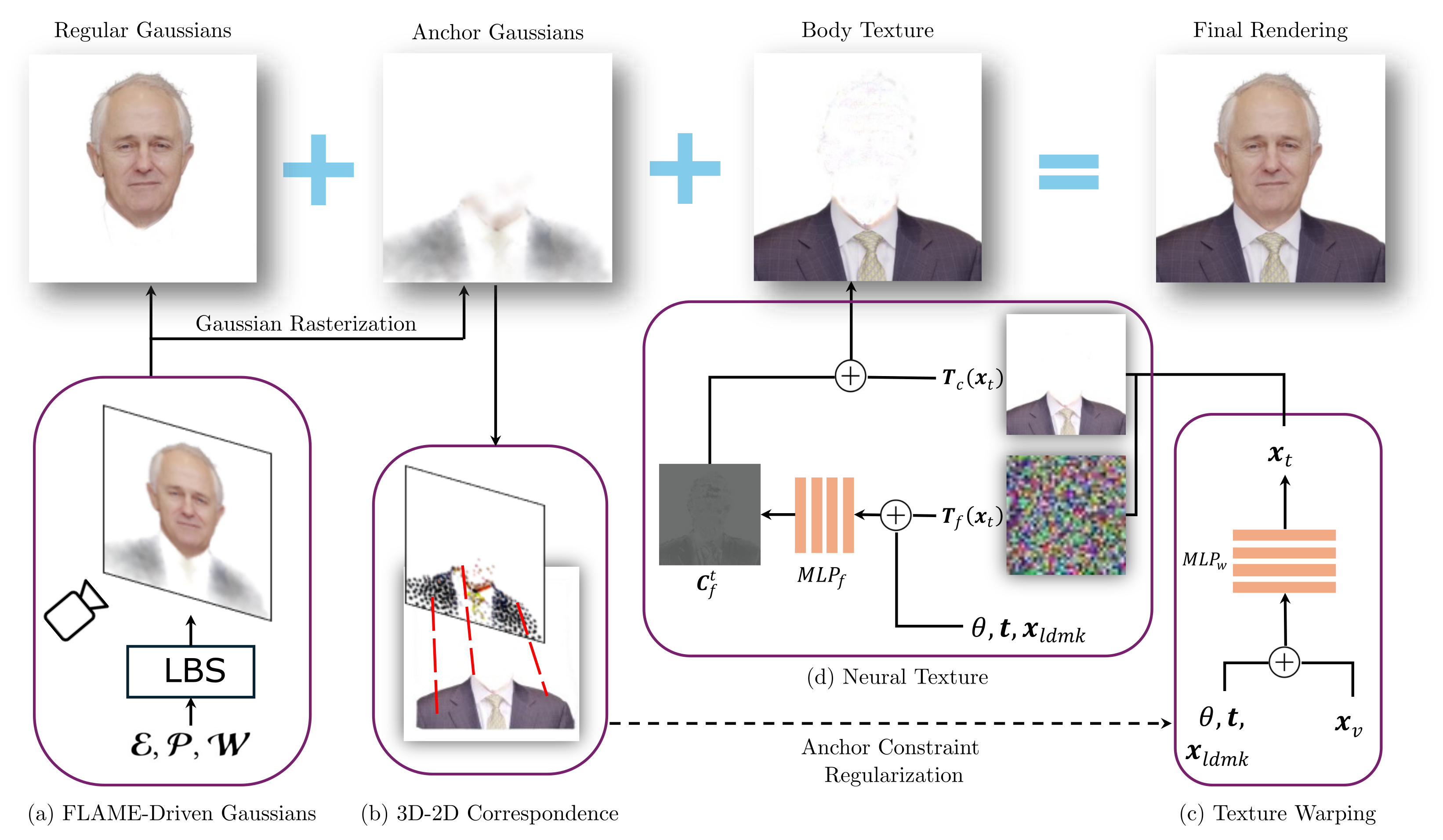}
    \caption{\textbf{Method.} (a) We utilize a set of standard head Gaussians and anchor Gaussians driven by LBS with the FLAME model. (b) Anchor Gaussians are initialized with a set of corresponding target coordinates in the texture space. This 3D-2D correspondence is used to constrain (c) a neural texture warping field that maps each pixel on the image plane $\mathbf{x}_v$ to a pixel in the texture space $\mathbf{x}_t$. (d) We then sample in the texture space to fetch the coarse texture $\mathbf{T}_c$ and latent texture $\mathbf{T}_f$, which is parsed by an MLP to obtain pose-dependent fine texture $\mathbf{C}_f^t$. Both coarse and fine textures are then combined to form a body texture, which is blended with other Gaussians through alpha compositing to form the final rendering.
    }
    \label{fig:method}
\end{figure*}

Given a monocular video featuring a talking subject with various expressions and head poses, our goal is to reconstruct a high-fidelity and animatable avatar including the head and clothed upper body. 
As illustrated in Fig~\ref{fig:method}, our method jointly optimizes 1) a set of standard 3D Gaussians \cite{gaussian_splatting} which tightly follow the transformation of 3DMM via LBS to represent the head region, 2) a set of sparse anchor Gaussians spawning over the clothed body, and 3) a learnable neural texture with pose-dependent neural texture warping field constrained by the anchor Gaussians to represent the clothed body with sharp details and high robustness.

\subsection{Preliminary- Gaussian Splatting}

3D Gaussian Splatting is a volumetric representation that utilizes a dense set of anisotropic Gaussians with varying opacity and view-dependent radiance to represent 3D geometry and appearance. Each Gaussian is described with four parameters: position (Gaussian mean) $\bm{\mu}$, 3D covariance matrix $\bm{\Sigma}$, opacity $\alpha$ and Spherical Harmonic (SH) coefficients $\mathbf{SH}$ for computing view-dependent RGB color. For ease of optimization, the covariance matrix is further decomposed into a scaling matrix $\mathbf{S}$, stored as a scaling vector $\mathbf{s}$, and a rotation matrix $\mathbf{R}$, stored as a quaternion vector $\mathbf{q}$. The covariance matrix is obtained as: $\bm{\Sigma} = \mathbf{RS}\mathbf{S}^T\mathbf{R}^T$.

To render 3D Gaussians to RGB images, their means are projected onto 2D image plane with standard projective transformation, while the projected covariance matrix is obtained by $\bm{\Sigma}' = \bm{JW\Sigma W}^T\bm{J}^T$, where $\mathbf{W}$ is the world to camera transformation and $\mathbf{J}$ is the Jacobian approximating the projective transformation~\cite{ewa_volume_splatting}. The rendered RGB color at each pixel is then obtained through:
\begin{align}
    \mathbf{C}(\mathbf{x}) &= \sum_{i\in N} \mathbf{c}_i \alpha^*_i(\mathbf{x}) \prod_{j=1}^{i-1}(1-\alpha^*_j(\mathbf{x})),
    \label{eq:gs_render}
    \\
    \alpha^*_i(\mathbf{x}) &= \alpha_i \exp \left( -\frac{1}{2}(\mathbf{x} - \bm{\mu}'_i)^T \bm{\Sigma}'^{-1} (\mathbf{x} - \bm{\mu}'_i) \right),
\end{align}
\noindent
where $\mathbf{x}$ is the 2D pixel coordinate, $\mathbf{c}_i$ is the view-dependent RGB radiance of i-th Gaussian on the ray obtained from SH function, $\alpha_i$ and $\bm{\mu}'_i$ are the opacity and projected 2D mean of the i-th Gaussian respectively.


\subsection{FLAME-Driven Head Gaussians}
As the face region contains highly distinguishable characteristics and can be described accurately with parametric head 3DMM such as FLAME~\cite{FLAME:SiggraphAsia2017}, we directly utilize standard 3D Gaussians that are deformed with parametric 3DMM via neural LBS to represent the head part~\cite{pointavatar, psavatar}. Specifically, we learn personalized FLAME expression and pose blendshapes and LBS weights through a small 3D coordinate-based MLP for each Gaussian:
\begin{align}
    \bm{\mathcal{E}}, \bm{\mathcal{P}}, \bm{\mathcal{W}} = \MLP_d(\bm{\mu}),
\end{align}
\noindent
where $\bm{\mathcal{E}} \in \mathbb{R}^{n_e \times 3}$ are the expression blendshapes, $\bm{\mathcal{P}} \in \mathbb{R}^{n_p \times 9 \times 3}$ are the pose blendshapes, $\bm{\mathcal{W}} \in \mathbb{R}^{n_j}$ are the LBS weights corresponding to each of the $n_j$ bones.
Following~\cite{hu2023gauhuman}, 
we use the standard skinning function \textsc{LBS} to obtain the rotation $\bm{R}$ and translation $\bm{T}$ for each Gaussian, and apply them to get the Gaussian mean $\bm{\mu}^d$ and covariance $\bm{\Sigma}^d$ in the 3D view space:
\begin{align}
    \bm{R}, \bm{T} &= 
    \textsc{LBS}(\bm{B}_{\bm{\mathcal{P}}}
    (\theta; \bm{\mathcal{P}}) 
    + \bm{B}_{\bm{\mathcal{E}}}(\psi; \bm{\mathcal{E}}),
    \mathbf{J}(\psi), \theta, \bm{\mathcal{W}}
    ),
    \\
    \bm{\mu}^d &= \bm{R}\bm{\mu} + \bm{T},
    \\
    \bm{\Sigma}^d &= \bm{R}\bm{\Sigma}\bm{R}^T,
\end{align}
\noindent
where $\mathbf{J}$ is the joint regressor in FLAME, and $\bm{B}_{\bm{\mathcal{P}}}$ and $\bm{B}_{\bm{\mathcal{E}}}$ are linear combination of blendshapes based on per-frame coefficients $\theta$ and $\psi$ that control the head animation.
They can then be rendered with a standard Gaussian rasterization pipeline in Eq~\ref{eq:gs_render}.

\subsection{3D-2D Correspondence via Anchor Gaussians}\label{sec:anchor_gs}

3D Gaussian Splatting has shown promising performance and robustness in reconstructing 3D geometry and appearance from RGB images. However, they suffer from a significant constraint -- each individual Gaussian can only represent a spatially invariant color under a fixed viewing direction, hence a vast number of Gaussians is required to represent objects with detailed textures, regardless of the actual complexity of the geometry. 
%
A naive application of Gaussian Splatting therefore fails to capture the fine details of the upper body with complex textures and intricate deformation, and results in blurry details and floating artifacts under challenging poses.

We hence propose to learn a high-frequency texture in canonical texture space, and use a sparse set of Gaussians as anchors to guide the warping between texture space and image plane.
As such, 
we only need a small number of Gaussians and a texture with per-pose warping to represent a clothed body with arbitrarily complicated textures.
%
Since anchor Gaussians themselves do not need to exactly represent the high-frequency appearance, we can model them as a simplified version of regular Gaussians:
they only use view-independent RGB colors, are isotropic Gaussians with quaternion fixed at $(1,0,0,0)$, and are excluded from the density control and therefore are not split, cloned, or pruned.
To prevent them from becoming trivial in rendering, their opacity and size are clamped to be no smaller than $0.05$ and $0.0001$ respectively.

The anchor Gaussians are initialized as follows: after a short warm-up period that only trains plain Gaussian, we first reproject all Gaussian means onto the image plane of a canonical training frame, and filter out Gaussians that are located around the head region based on semantic masks.
We then use farthest point sampling~\cite{qi2017pointnetplusplus} to select $N_a=1024$ Gaussians as anchor Gaussians. The first SH basis is converted to RGB values and the anchor scales in three directions are averaged to form a single scale for the anchor Gaussians. 
We then obtain a sparse set of anchor Gaussians, as well as their projected 2D means $\hat{\mathbf{x}}^v_i$ on the image plane (2D view space) of the canonical frame:
%
%
\begin{align}
    \hat{\mathbf{x}}^v_i &= \mathbf{P}(\hat{\bm{\mu}}^d_i) ,
    \label{eq:anchor_projection_view_space}
\end{align}
\noindent
where 
$\mathbf{P}$ is the camera projective transformation, $\hat{\bm{\mu}}_i^d$ is the 3D Gaussian mean of the $i$-th anchor Gaussian transformed to 3D view space with LBS. 
%
To build the correspondence between anchor Gaussians and texture space coordinates, we assume that the mapping between the 2D image plane of the canonical frame and the texture space is an identity mapping. We can hence define a function $f_{anchor}(i)$ as a fixed correspondence between the $i$-th 3D anchor Gaussian mean and its target 2D pixel coordinate in texture space:
\begin{align}
    f_{anchor}(i) := \mathbf{I}(\hat{\mathbf{x}}^v_i),
\end{align}
\noindent
where $\mathbf{I}$ is the identity function to map 2D image plane coordinates to texture space. Note that $f_{anchor}(i)$ is fixed after initialization and does not update with further optimization of $\hat{\bm{\mu}}_i$. Such correspondences will later be used to constrain the pose-dependent texture warping, as will be detailed in Sec~\ref{sec:optimization}.
%



\subsection{Neural Texture and Texture Warping}


We use a trainable neural texture in canonical space with a pose-dependent neural texture warping field to represent the part of the avatar with relatively simple overall geometry and complicated appearances, i.e., the clothed shoulder and chest. 
In a traditional textured mesh rendering pipeline, the texture is first mapped to the mesh triangles through a pre-defined UV mapping, and the meshes are then rasterized to find the first intersections with the camera rays. Those first intersections therefore establish a mapping between texture space and image plane. However, this approach is not applicable without accurate surfaces and well-defined UV mapping. We instead propose to bypass the intermediate step and learn a per-pose warping that \ww{directly maps pixel coordinates on image plane $\mathbf{x}_v$ to the texture coordinates $\mathbf{x}_t$ for texture fetching}. Specifically, the warping field is represented using a coordinate-based MLP: 
\begin{align}
    \Delta_\mathbf{x} = \MLP_w \left(  \gamma(\mathbf{x}_v), \gamma(\theta), \gamma(\mathbf{t}), \gamma(\mathbf{x}_{ldmk}) \right) ,
\end{align}
\noindent
where
$\gamma$ is the positional encoding~\cite{nerf}, 
$\mathbf{\theta}$ is the FLAME pose parameters containing head and neck rotations, $\mathbf{t}$ is the camera position, $\mathbf{x}_{ldmk}$ is 2D body landmarks for neck, left and right shoulders. 
 The corresponding texture coordinate is obtained as $\mathbf{x}_t = \mathbf{x}_v + \Delta_\mathbf{x}$. 
 
Our optimizable texture includes a coarse texture $\mathbf{T}_c$ with 3 channels and a latent texture $\mathbf{T}_f$ with $D_t$ channels. Both textures have sizes of $[H+2P, W+2P]$, where $H, W$ are the image height and width, $P$ is the padding size which we empirically set to 50 to account for body parts that move in and out in the video sequence. 
The latent texture $\mathbf{T}_f$ is passed to an MLP to obtain pose-dependent appearances such as brightness changes on the clothes:
\begin{align}
    \mathbf{C}^t_f(\mathbf{x}_t) &= \MLP_f \left( \mathbf{T}_f(\mathbf{x}_t), \gamma(\theta), \gamma(\mathbf{t}), \gamma(\mathbf{x}_{ldmk}) \right) ,
\end{align}
\noindent
where $ \mathbf{T}_c(\mathbf{x}_t),  \mathbf{T}_f(\mathbf{x}_t)$ are coarse and latent texture sampled at 2D coordinate $\mathbf{x}_t$ via bilinear interpolation.
%
The textured pixel color at the coordinate $\mathbf{x}_v$ is therefore obtained as:
 \begin{align}
    \mathbf{C}^t(\mathbf{x}_v) &= \mathbf{T}_c(\mathbf{x}_t) + \mathbf{C}^t_f(\mathbf{x}_t) .
 \end{align}

\subsection{Rendering}

To this end, we have a hybrid representation that includes 3D regular Gaussians that represent the head of the avatar, 3D anchor Gaussians that sparsely span over the body region, and a 2D neural texture for the body. To render all of them together for joint optimization, we simply use alpha blending:
\begin{equation}
\begin{aligned}
    \mathbf{C}^*(\mathbf{x}_v) &= 
    \underbrace{\hat{\mathbf{C}}(\mathbf{x}_v)}_{\text{Anchor Gaussians}} +\underbrace{(1 - \hat{\alpha}(\mathbf{x}_v)) \mathbf{C}(\mathbf{x}_v)}_{\text{Head Gaussians}}
    \\
    &+\underbrace{(1 - \hat{\alpha}(\mathbf{x}_v)) (1 - \alpha(\mathbf{x}_v)) \mathbf{C}^t(\mathbf{x}_v)}_{\text{Body Texture}} ,
    \label{eq:render}
\end{aligned}
\end{equation}
\noindent
where $\hat{\mathbf{C}}(\mathbf{x}_v),  \mathbf{C}(\mathbf{x}_v)$ are the rendered RGB color of anchor Gaussian and regular Gaussian, $\hat{\alpha}(\mathbf{x}_v),  \alpha(\mathbf{x}_v)$ are the total alpha of anchor Gaussian and regular Gaussian at pixel $\mathbf{x}_v$ respectively.

Note that our rendering process always renders anchor Gaussians in front of the regular Gaussians regardless of their actual positions.
Though not physically realistic, 
we designed this rendering order so anchor Gaussians are always non-trivial and never occluded by regular Gaussians. 

\subsection{Optimization}
\label{sec:optimization}


The optimization is split into three different stages: anchor warm-up stage, main optimization stage, and texture refinement stage. In the anchor warm-up stage, neither anchor Gaussians nor body texture is applied, only the regular Gaussians are rendered and optimized.
The purpose of this stage is to move Gaussians to 
roughly spawn over 
the area of interest including both head and body.
At the end of this stage, we initialize anchor Gaussians from regular Gaussians using the method described in Sec~\ref{sec:anchor_gs}.
In the second stage, we render all of the regular Gaussians, anchor Gaussians, and the textured body with alpha compositing described in Eq~\ref{eq:render} and jointly optimize them together. In the last stage, to recover faithful appearance for the body texture and enhance its robustness under novel poses, we remove anchor Gaussians from the rendering pipeline, i.e., we set $\hat{\mathbf{C}}$ and $\hat{\alpha}$ to $0$ in Eq~\ref{eq:render}., and freeze everything else except for the neural texture, texture warping field, and opacity and SH of regular Gaussians. 

Following ~\cite{pointavatar, zheng2022imavatar}, the training losses include standard MSE RGB loss $\mathcal{L}_{\bm{C}} = MSE(\mathbf{C}^* - \mathbf{C}^{GT})$, and a FLAME regularization that encourages the FLAME blendshapes and LBS weights predicted for each Gaussian stay close to the pseudo ground truth $\Tilde{\bm{\mathcal{E}}}_i, \Tilde{\bm{\mathcal{P}}}_i, \Tilde{\bm{\mathcal{W}}}_i$ obtained from the nearest FLAME vertex:
\begin{equation}
\begin{aligned}
    \mathcal{L}_{flame} &= \frac{1}{N}\sum_{i=1}^{N+N_a} (
    \lambda_{\bm{\mathcal{E}}} |\bm{\mathcal{E}}_i - \Tilde{\bm{\mathcal{E}}}_i |_2
    \\
    &+ \lambda_{\bm{\mathcal{P}}} |\bm{\mathcal{P}}_i - \Tilde{\bm{\mathcal{P}}}_i |_2
    + \lambda_{\bm{\mathcal{W}}} |\bm{\mathcal{W}}_i - \Tilde{\bm{\mathcal{W}}}_i |_2
    ).
\end{aligned}
\end{equation}
%
During main optimization stage, we additionally include a VGG feature loss~\cite{vgg_loss, vgg} $\mathcal{L}_{VGG} = | \mathbf{F}_{vgg}(\mathbf{C}) -\mathbf{F}_{vgg}(\mathbf{C}^{GT})|$, and a head mask regularization to encourage regular Gaussians to stay only within the head region and allow the body texture to be trained properly without being occluded:
\begin{align}
    \mathcal{L}_{head} = MSE(max(\alpha - \alpha_{head}, 0)) ,
\end{align}
\noindent
where $\alpha_{head}$ is the alpha mask of the head region obtained with matting pre-processing and semantic mask. 
We also include an L1 regularization on the 2D neural warping field to encourage a clean background to be learned in the neural texture, as well as an L1 loss to slowly decrease the opacity of anchor Gaussians to allow the body texture to be trained properly: 
\begin{align}
    \mathcal{L}_{warp} &= \frac{1}{HW} \sum_{i=1}^{HW} |\Delta_{\mathbf{x}_i}|,
    \\
    \mathcal{L}_{\hat{\alpha}} &= \frac{1}{N_a} \sum_{i=1}^{N_a} |\hat{\alpha}_i|.
\end{align}
Finally, we include an anchor loss as a soft constraint of the per-pose texture warping: 
\begin{align}
    \mathcal{L}_{anchor} &= \frac{1}{N_a} \sum_{i=1}^{N_a}(f_{anchor}(i) -  (\hat{\bm{x}}^v_i + \Delta_{\hat{\bm{x}}^v_i} ))^2,
\end{align}
\noindent
i.e., for each anchor Gaussian, we first transform it to 3D view space via LBS, and then project it onto the image plane to obtain its 2D mean $\hat{\bm{x}}^v_i$ via Eq~\ref{eq:anchor_projection_view_space}. $\hat{\bm{x}}^v_i$ is then warped by the neural warping field $\MLP_w$ to obtain the corresponding coordinate in the texture space, which is optimized to match the anchor correspondence defined during initialization.

In the third stage, we remove the regularization losses including $\mathcal{L}_{head}, \mathcal{L}_{warp}$ and $\mathcal{L}_{\hat{\alpha}}$.

The total training objectives for each of the three stages are as follows:
\begin{align}
\mathcal{L}_1 &= \mathcal{L}_{\bm{C}}  + 
    \mathcal{L}_{flame},
    \\
\mathcal{L}_2 &=
    \mathcal{L}_1
    + \lambda_{VGG}\mathcal{L}_{VGG} 
    + \lambda_{head}\mathcal{L}_{head} \nonumber
    \\
    &\quad + \lambda_{warp}\mathcal{L}_{warp}
    + \lambda_{\hat{\alpha}}\mathcal{L}_{\hat{\alpha}}
    + \lambda_{anchor}\mathcal{L}_{anchor},
\\
 \mathcal{L}_3 &= \mathcal{L}_1 
 + \lambda_{VGG}\mathcal{L}_{VGG} 
 + \lambda_{anchor}\mathcal{L}_{anchor}.
\end{align}

\subsection{Accelerated Rendering with No MLP Queries}

One of the main advantages of Gaussian Splatting is its highly efficient rendering speed, which enables many real-time and interactive applications. To take full use of this advantage, we propose an accelerated version of our method that requires no MLP queries at inference time. Specifically, after training the model, we first cache the output of $\MLP_d$ for all head Gaussians and anchor Gaussians, then cache the view-dependent fine texture by querying the fine texture MLP $\MLP_f$ conditioned on the same canonical training frame which was previously used to initialize the anchor Gaussians. The queried fine texture colors are added to the coarse color to make a non-neural RGB texture. To deal with potential noise created by the fine texture MLP at the corners of the texture, we use an off-the-shelf background segmentation network~\cite{deeplabv3} to compute a coarse mask and clean all the pixels outside of the mask; \spl{we show the necessity of this step in the supplementary}.
To replace the neural warping field $\MLP_w$ that warps image plane coordinates to texture space, we rely on the correspondence between anchor Gaussians and texture space coordinates to estimate a homography at inference time. 
Specifically, we first project all anchor Gaussians to the image plane of the canonical training frame, and then remove any anchor Gausians that go beyond the view frustum. 
To deal with any potential discrepancy between the neural warping field and the anchor correspondences, we update those correspondences based on the prediction of the neural warping field on the current frame:
\begin{align}
    f_{anchor}(i) := \hat{\mathbf{x}}_{v}^i + \Delta_{\hat{\mathbf{x}}_{v}^i}.
\end{align}
After that, we randomly select 100 training frames and use RANSAC \cite{ransac} to estimate a homography between the image plane coordinates of anchor Gaussians and their corresponding texture space coordinates, and remove anchor Gaussians that are considered outliers by RANSAC. This effectively removes any anchor deformation that cannot be described by the rigid transformation.
Finally, at inference time, we perform LBS on regular head Gaussians and anchor Gaussians. Based on the image plane coordinates of the anchor Gaussians $\hat{\mathbf{x}}_{v}^i$ and their correspondences $f_{anchor}$, we compute a homography with the least square error via singular value decomposition. The estimated transformation is applied to all pixels on the image plane to find the corresponding non-neural texture, which is then blended with the head Gaussians to form the final rendering. 
This accelerated inference approach effectively increases the rendering speed from around 70 FPS to 130 FPS.


\section{Evaluation}

\begin{table*}[t]
\small
\centering
\singlespacing
\tabcolsep=0.06cm
\begin{NiceTabular}{l | lll | lll | lll | lll | lll | lll | lll}
\toprule
\multicolumn{1}{c}{}  & \multicolumn{3}{c|}{001}  & \multicolumn{3}{c|}{002} & \multicolumn{3}{c|}{003} & \multicolumn{3}{c|}{004} & \multicolumn{3}{c}{005} & \multicolumn{3}{c}{006} & \multicolumn{3}{c}{007}\\
{} &        \scalebox{0.8}{PSNR} &      \scalebox{0.8}{SSIM} &      \scalebox{0.8}{LPIPS} &        \scalebox{0.8}{PSNR} &       \scalebox{0.8}{SSIM} &      \scalebox{0.8}{LPIPS} &        \scalebox{0.8}{PSNR} &       \scalebox{0.8}{SSIM} &      \scalebox{0.8}{LPIPS} &        \scalebox{0.8}{PSNR} &       \scalebox{0.8}{SSIM} &      \scalebox{0.8}{LPIPS} &        \scalebox{0.8}{PSNR} &       \scalebox{0.8}{SSIM} &      \scalebox{0.8}{LPIPS} &        \scalebox{0.8}{PSNR} &       \scalebox{0.8}{SSIM} &      \scalebox{0.8}{LPIPS} &        \scalebox{0.8}{PSNR} &       \scalebox{0.8}{SSIM} &      \scalebox{0.8}{LPIPS} \\
\midrule
INSTA           &           18.58 &           .751 &           .269 &           22.90 &           .880 &          .177 &           22.24 &           .809 &           .175 &           19.45 &           .784 &          .310 &           19.47 &           .757 &           .251 &           23.44 &           .861 &           .165 &           18.68 &           .733 &           .291 \\
SplattingAvatar &           18.58 &           .738 &           .300 &           25.34 &           .876 &          .171 &           21.34 &           .790 &           .220 &           19.83 &           .765 &          .351 &           20.06 &           .763 &           .250 &           22.78 &           .838 &           .201 &           20.15 &           .754 &           .257 \\
PointAvatar     &           22.83 &           .822 &           .100 &           30.61 &           .924 &          .062 &           28.12 &           .874 &           .077 &           23.99 &  \second{.837} &          .133 &  \second{22.82} &           .847 &           .142 &           29.42 &  \second{.929} &           .043 &           22.30 &           .826 &           .088 \\
GS*             &           23.26 &           .814 &           .082 &   \first{32.99} &           .937 &          .046 &  \second{29.85} &           .888 &           .054 &           24.18 &           .836 &          .139 &           22.80 &           .847 &           .129 &  \second{29.56} &           .924 &           .039 &  \second{22.31} &           .820 &           .099 \\
Ours            &   \first{25.95} &   \first{.856} &   \first{.064} &  \second{31.98} &   \first{.949} &  \first{.042} &   \first{31.26} &   \first{.917} &   \first{.042} &   \first{24.68} &   \first{.839} &  \first{.120} &   \first{24.48} &   \first{.895} &   \first{.074} &   \first{30.97} &   \first{.943} &   \first{.033} &   \first{23.26} &   \first{.856} &   \first{.074} \\
Ours (No MLP)     &  \second{24.48} &  \second{.840} &  \second{.070} &           31.44 &  \second{.942} &  \first{.042} &           28.85 &  \second{.892} &  \second{.051} &  \second{24.61} &  \second{.837} &  \first{.120} &           22.19 &  \second{.860} &  \second{.078} &           28.71 &           .912 &  \second{.037} &           21.49 &  \second{.827} &  \second{.081} \\
\bottomrule
\end{NiceTabular}
    \caption{\textbf{Quatitative evaluation of self-reenactment task} We report PSNR$\uparrow$, SSIM$\uparrow$, and LPIPS$\downarrow$ ,and color the \colorbox{first}{best} and \colorbox{second}{second-best} methods for each subject respectively. Our full method achieves much better performance compared to existing baselines. While Ours (No MLP) sometimes achieves slightly lower PSNR, which is known to be over-sensitive to small misalignments and prefer blurry results \cite{park2021hypernerf}, we show it achieves better LPIPS than existing methods.
    }
    \label{tab:self_reenact}
\end{table*}

\begin{table}[t]
\small
\centering
\singlespacing
\tabcolsep=0.06cm
\begin{NiceTabular}{l | cc | cc | cc | cc}
\toprule
\multicolumn{1}{c}{}  & \multicolumn{2}{c|}{003} & \multicolumn{2}{c|}{004}  & \multicolumn{2}{c|}{005} & \multicolumn{2}{c}{007}
\\
{} & FPS & \#GS & FPS & \#GS & FPS & \#GS  & FPS & \#GS 
\\
\midrule
GS*  & \textbf{141} & 163830 & \textbf{159} & 125521 & 96 & 317968 & \textbf{131} & 191431
\\
Ours  & 70 & 58701 & 71 & 39549 & 72 & 50708 & 69 & 52910
\\
Ours (No MLP)  & 129 & 58701 & 134 & 39549 & \textbf{132} & 50708 & 127 & 52910
\\
\bottomrule
\end{NiceTabular}
    \caption{\textbf{Performance measure.} We report rendering FPS and the number of Gausssians for each method. 
    }
    \label{tab:fps}
\vspace{-10pt}
\end{table}


\subsection{Datasets} 
We evaluate different methods on 1 mobile phone sequence from PointAvatar~\cite{pointavatar}, 2 internet sequences from Head2Head dataset~\cite{head2head2020}, and 4 sequences captured with mobile phones. All sequences are preprocessed with DECA~\cite{DECA:Siggraph2021} and a slightly modified landmark fitting process from IMAvatar~\cite{zheng2022imavatar}. Additionally, we use DWPose~\cite{dwpose} to predict 2D landmarks for nose, neck and shoulders, which are then smoothed with One Euro Filter~\cite{oneeurofilter}.

\subsection{Baselines}
We compare our method with four neural head avatar methods based on various representations, including (1) INSTA~\cite{INSTA}, which employs a latent hash grid~\cite{instantNGP} combined with NeRF~\cite{nerf}, (2) PointAvatar~\cite{pointavatar}, which is based on isotropic point clouds, (3) SplattingAvatar~\cite{SplattingAvatar:CVPR2024}, which utilizes Gaussian Splatting attached to local space of 3DMM meshes, and (4) GS*, a baseline we implemented by changing the point cloud representation in PointAvatar to Gaussian Splatting, which is similarly deformed via neural LBS.

\subsection{Self-Reenactment}
\begin{figure*}
    \centering
    \includegraphics[width=0.9\textwidth]{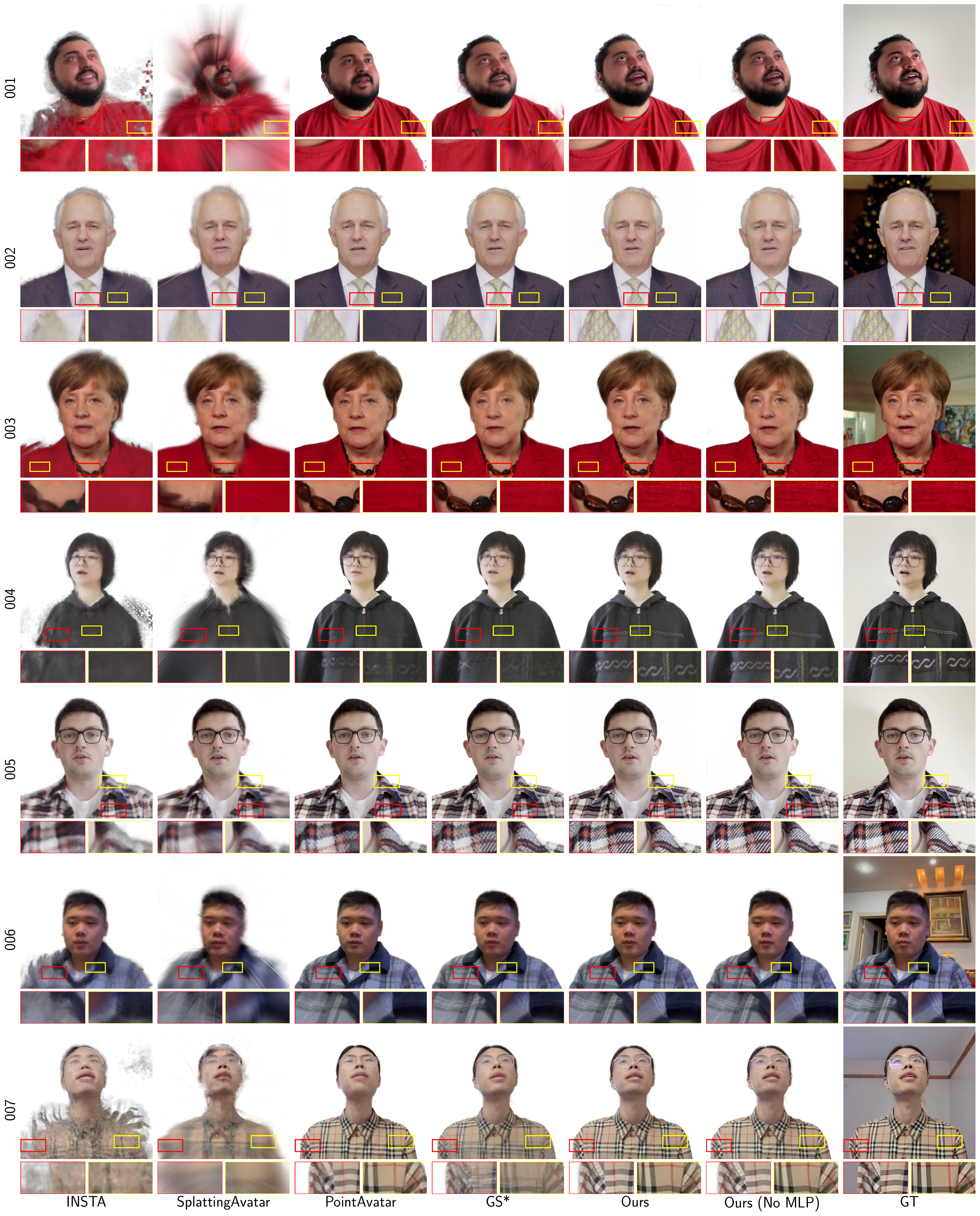}
    \caption{\textbf{Qualitative comparison of self-reenactment task.} We show that both full version and No MLP version of our method can effectively recover a more accurate and more robust body texture, even under cases of extreme poses and high-frequency cloth textures.}
    \label{fig:main}
\end{figure*}

We show the quantitative and qualitative results of the self-reenactment task in Tab~\ref{tab:self_reenact} and Fig~\ref{fig:main}. The subjects 001 to 007 are shown in order in Fig~\ref{fig:main}.
Our full version demonstrates superior reconstruction performance compared to existing baselines, especially for subjects with intricate cloth textures. 
Our No MLP version does not consistently achieve better PSNR when compared to existing baselines, as it is unable to render pose-dependent appearance changes and intricate cloth deformation. However, we note that it consistently achieves better LPIPS, demonstrating that our No MLP version can still generate realistic and faithful renderings.
This discrepancy among different metrics arises because of the high sensitivity of PSNR to small misalignments in the cloth texture~\cite{park2021hypernerf}. 
As a result, PSNR tends to prefer blurry reconstruction over sharp but slightly misaligned results. 
The qualitative evaluation in Fig~\ref{fig:main} demonstrates that both versions of our method are capable of learning sharper and more robust body texture compared to existing methods. 
Specifically, INSTA~\cite{INSTA} and SplattingAvatar~\cite{SplattingAvatar:CVPR2024} fail to learn sensible reconstruction for the body part, as they originally only aim for the reconstruction of head and neck without any body; PointAvatar~\cite{pointavatar} learns robust head and body avatars, but produces highly blurred results; GS* learns sensible reconstruction with accurate expression and pose control, but it still misses some sharp textures and can cause floating artifacts and blurry reconstruction under extreme poses. In comparison, our methods can produce much sharper and more reliable rendering under any texture and pose. Please refer to the video results in the Supplementary for additional comparisons.

\subsection{Cross-Reenactment}
For the cross-identity reenactment task, we render the reconstruction of the original identity with FLAME expressions and poses from the source subject. As the subjects in different videos have various crop sizes and distances to the camera, we do not directly take the body landmarks of source subjects, but instead apply the offsets between current landmarks and the landmarks at the canonical frame of the source subject to the landmarks of original identity. 
With the full version of our method, we apply an additional Euclidean transformation after warping the image plane coordinates with the MLP. 
This is needed to ensure the body texture is always aligned with the head Gaussians under novel poses; see Fig~\ref{fig:abla_reenact}.
The Euclidean transformation is simply determined by fitting the MLP warped image plane coordinates of the anchor Gaussians and their target coordinates in the texture space. 
To deal with potential artifacts caused by coordinates warped to unseen corner parts in the texture, we apply the same appearance distillation process and remove the fine texture MLP. 
The No MLP version is applied the same way as in the self-reenactment task.

\begin{figure*}
    \centering
    \includegraphics[width=0.8\textwidth]{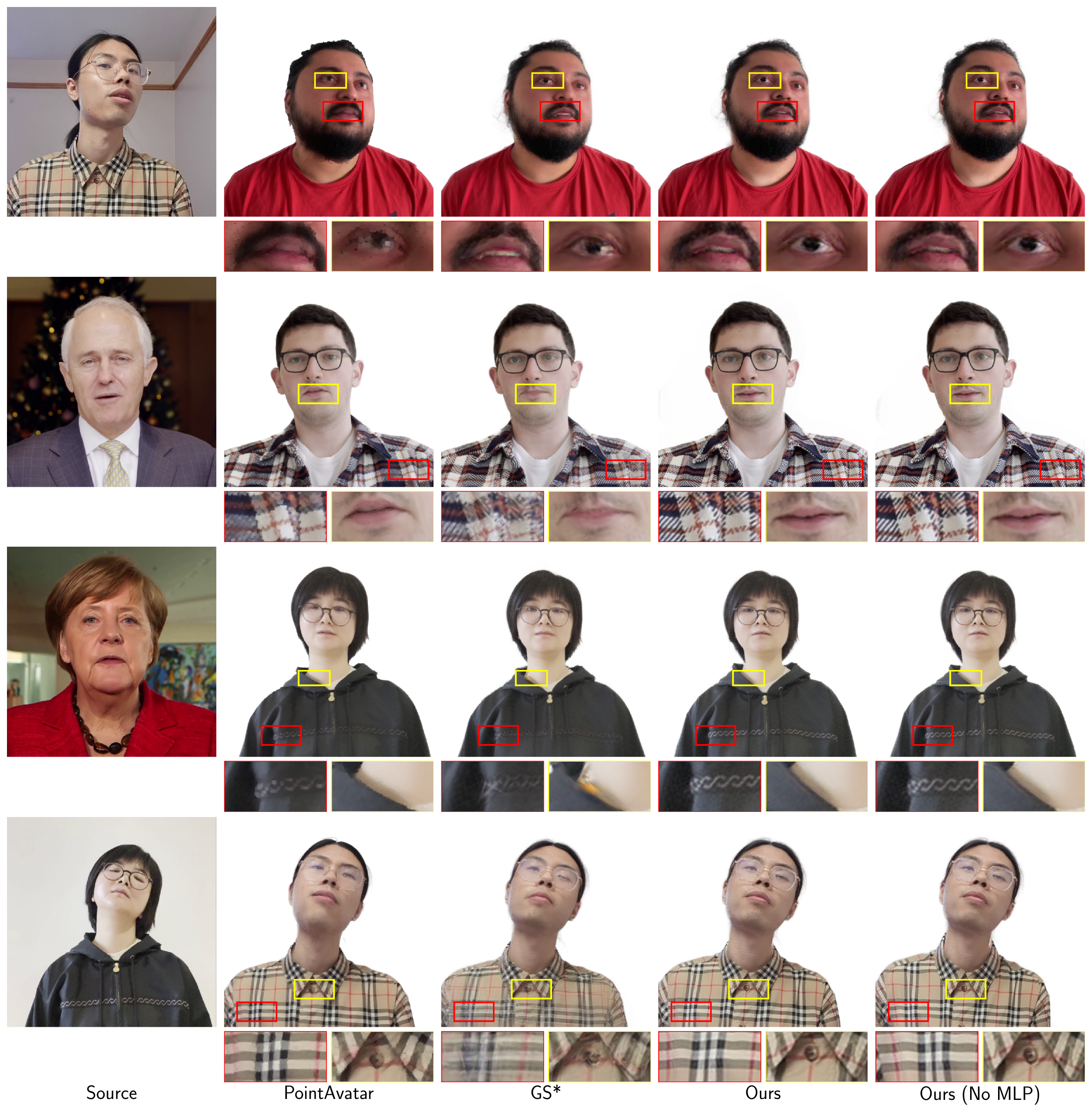}
    \caption{\textbf{Qualitative evaluation of cross-identity reenactment.} In addition to the improvement in cloth texture quality and robustness, we found that our approach often leads to more accurate expression control. This is because we are using much fewer LBS-driven Gaussians for the body part, therefore the capacity of LBS weight inference network can fully focus on the head region. }
    \label{fig:cross_reenact}
\end{figure*}

In addition to the improvement over the body texture, we observe that avatars reconstructed with our approach often give more accurate and faithful expression control, as shown in Fig~\ref{fig:cross_reenact}. 
We deduce that this is because the 3DMM-driven Gaussians only need to model the head region, leading to a more accurate reconstruction of the head model and more reliable LBS weights and expression and pose blendshapes predicted by the LBS network.

\subsection{Ablation}

\begin{figure}
    \centering
    \includegraphics[width=0.375\textwidth]{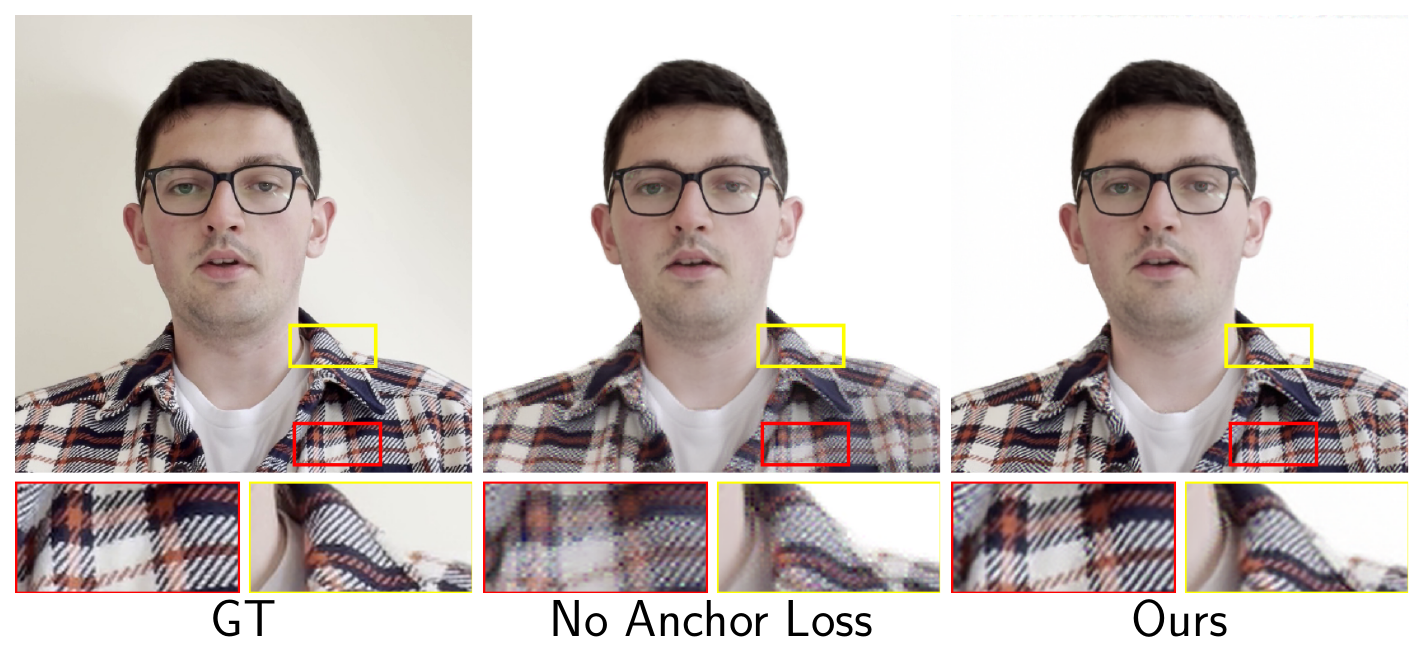}
    \caption{\textbf{Ablation.} The anchor constraint is necessary for learning sharp textures even if the body only moves slightly during the video.}
    \label{fig:abla}
\end{figure}

\begin{figure}
    \centering
    \includegraphics[width=0.475\textwidth]{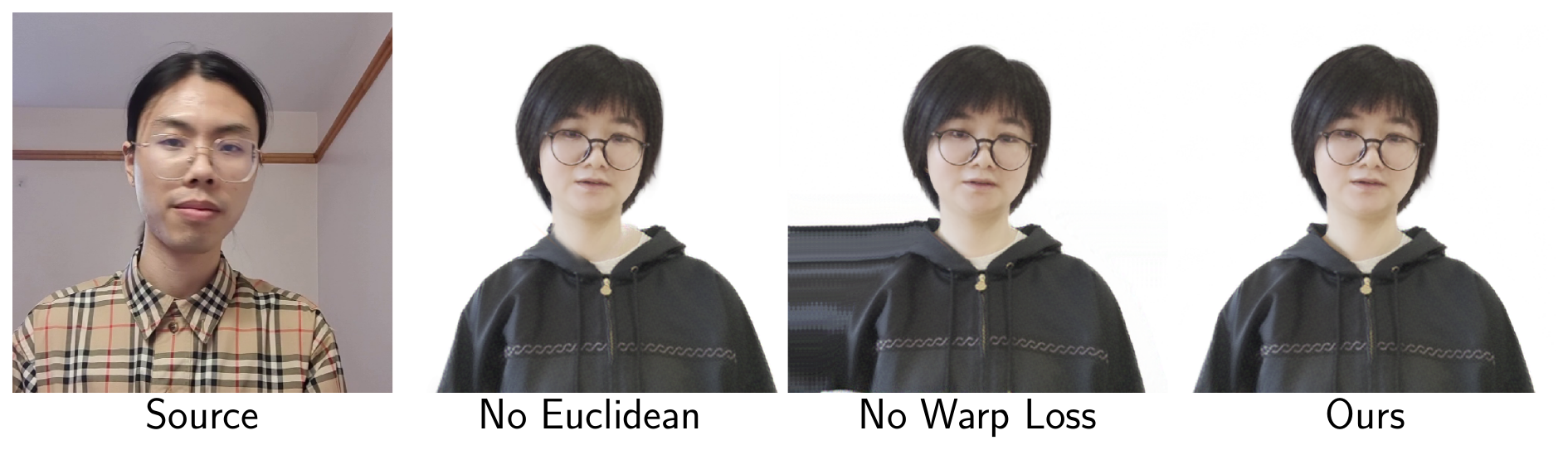}
    \caption{\textbf{Ablation for cross-identity reenactment.} The additional Euclidean transformation helps to align the body texture with head Gaussian under novel poses, whereas $\mathcal{L}_{warp}$ is necessary to prevent arbitrary warping of the white background.}
    \label{fig:abla_reenact}
\end{figure}

We show the effectiveness of the anchor constraint $\mathcal{L}_{anchor}$, test-time Euclidean transformation and warp loss $\mathcal{L}_{warp}$ in Fig~\ref{fig:abla} and~\ref{fig:abla_reenact}. Even for subjects with only slight movement in the upper body, anchor constraint is still needed to learn sharp and accurate cloth texture. Besides, without anchor Gaussians and test time Euclidean transformation, the body texture is unable to align with the head Gaussians under novel poses. The warp loss $\mathcal{L}_{warp}$ is needed to prevent the neural warping field from mapping the background pixel to an arbitrary white pixel in the texture space. As anchor Gaussians only exist within the body region, the additional Euclidean transformation computed from anchor correspondences would significantly distort the background pixels, causing severe artifacts as shown in Fig~\ref{fig:abla_reenact}. \spl{Additional ablation results can be found in the supplementary}.

\subsection{Rendering Efficiency}

We report the number of Gaussians and the rendering speed for pure Gaussian implementation GS*, Ours, and Ours (No MLP) in Tab~\ref{tab:fps}. The rendering speeds are tested on an RTX4080 Ti. For subjects wearing complicated clothes, the number of Gaussians required to model the high-frequency cloth texture significantly increases for pure Gaussian implementation, hence slowing down the rendering speed, whereas our method only models the head region with Gaussians and hence requires a much fewer number of Gaussians. The rendering speed of our no MLP version even surpasses pure Gaussian implementation for subject 005, who wears cloth with a very high-frequency texture.

\section{Conclusion}

We present Gaussian Head \& Shoulders,
a method that reconstructs high-quality and animatable upper body avatars including head, chest and shoulders. By utilizing high-frequency neural texture to represent the clothed body, we are able to model sharp and robust cloth details and significantly reduce the number of Gaussians needed to represent a subject. By constraining the texture warping with a sparse set of anchor Gaussians, the body texture
is accurately mapped to the correct position even under unseen poses. 
By caching the neural texture and replacing the neural warping field with a projective transformation estimated using anchor correspondences, we significantly improve rendering speed and reach over 130 FPS at novel poses, surpassing the rendering speed of pure Gaussian implementation for subjects with complicated cloth textures. 

\paragraph{Limitation} 
Our approach cannot model avatars with extreme body rotation that causes self-occlusion in the video, as our neural texture warping does not account for self-occlusion. However, such extreme rotations are rare for common applications such as video conferencing. More discussion can be found in the supplementary.

\clearpage

\appendix
{
    \centering
    \Large
    \textbf{Gaussian Head \& Shoulders:  High Fidelity Neural Upper Body Avatars with Anchor Gaussian Guided Texture Warping} \\
    \vspace{0.5em}Supplementary Material \\
    \vspace{1.0em}
}


In this supplementary material, we provide additional implementation and evaluation details in Sec~\ref{sec:implementation_details}, as well as extended results including additional ablation studies, limitations, and a comparison with SMPL-driven body avatar in Sec~\ref{sec:additional_results}. Ethic discussions are in Sec~\ref{sec:ethics}. We also highly recommend the readers to view our supplementary video.

\section{Implementation Details}\label{sec:implementation_details}

\subsection{Preprocessing}
\begin{figure}
    \centering
    \includegraphics[width=0.45\textwidth]{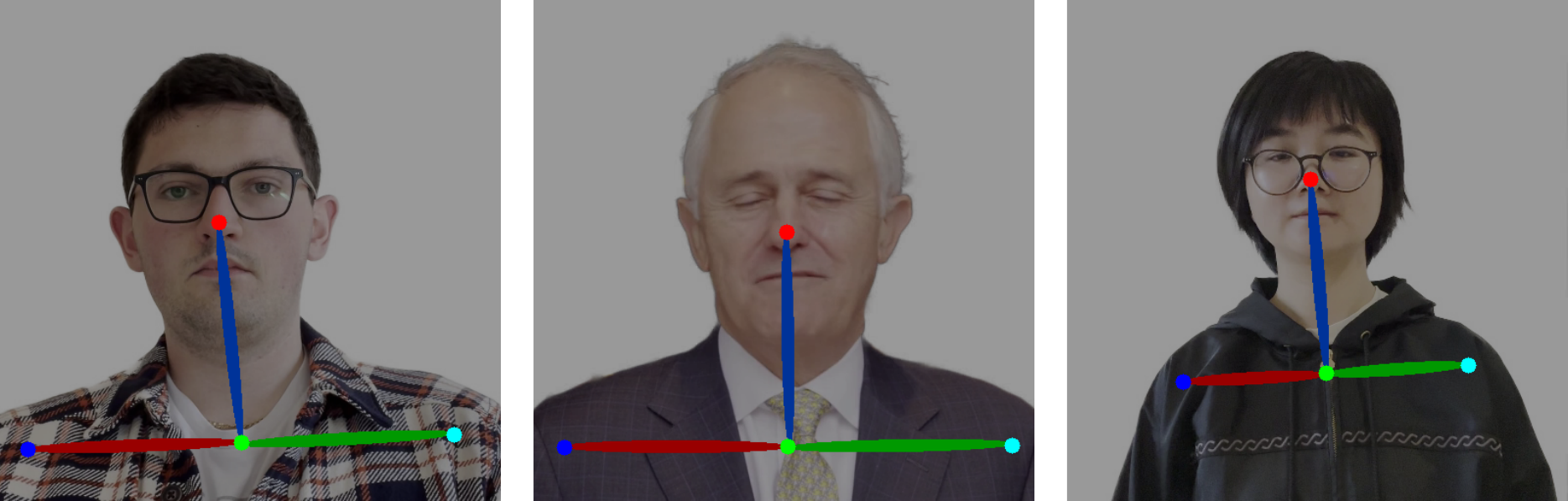}
    \caption{\textbf{Landmarks.} We use DWPose~\cite{dwpose} to detect nose, neck and shoulder landmarks to use as input to $\MLP_f$ and $\MLP_w$.}
    \label{fig:ldmk}
\end{figure}

Our data preprocessing pipeline for extracting FLAME parameters, camera parameters and body landmarks is modified from~\cite{zheng2022imavatar}. After obtaining rough FLAME parameters from DECA~\cite{DECA:Siggraph2021}, we further optimize the FLAME parameters to minimize the 68 facial landmarks for 3000 iterations. For subject 001, we keep the original training and test split used by PointAvatar~\cite{pointavatar}. For other subjects, we use the last 500 or 1000 frames as test frames, depending on the total frame count in the video.
For all subjects, we simply use the first frame as the canonical training frame for initializing anchor Gaussians and updating the anchor correspondences. 
We use DWpose~\cite{dwpose} to detach the noise, neck and shoulder landmarks, which are illustrated in Fig~\ref{fig:ldmk}.

\subsection{Network Architecture}
We have three MLPs in total: $\MLP_d$ which predicts the expression blendshapes  $\bm{\mathcal{E}}$, pose blendshapes $\bm{\mathcal{P}}$ and LBS weights $\bm{\mathcal{W}}$ for each regular Gaussian and anchor Gaussian; $\MLP_f$ which predicts pose-dependent fine texture; $\MLP_w$ which warps view space coordinates to texture space coordinates. All three MLPs have 4 hidden layers and 128 neurons in each hidden layer. The standard Fourier frequency positional encoding~\cite{nerf} is applied to the pixel coordinate, FLAME head rotation, camera translation and 2D landmarks before inputting to $\MLP_f$ and $\MLP_w$. The pixel coordinate and 2D landmarks are encoded with a frequency of 10, and camera translation and FLAME head rotation are encoded with a frequency of 2. All three MLPs are initialized to predict 0s at the beginning by setting the weights and bias of the output layer to 0. All MLPs use ReLU as the intermediate activations. Tanh is used as the final activation for $\MLP_f$, no final activation is used for $\MLP_w$, and the final activation for $\MLP_d$ are the same as~\cite{pointavatar}. 

We use a latent dimension $D_t = 32$ for the latent texture $\mathbf{T}_f$. The coarse texture $\mathbf{T}_c$ is initialized to be the same as the white background, while the fine latent $\mathbf{T}_f$ is initialized and a random and uniform distribution between $[0,1]$.

\subsection{Training Details}

For all subjects, we use $\lambda_{head} = 1$, $\lambda_{anchor} = 1$, $\lambda_{warp} = 0.025$, $\lambda_{\hat{\alpha}} = 0.15$. For VGG loss weight $\lambda_{VGG}$, we set it to 0 for the first 10K iterations, and then $0.1$ for the rest of the training. This is needed as we empirically observe that training the neural texture and warping field with a strong VGG loss from the beginning severely harms their stability. The weights of FLAME regularization are initially set to $\lambda_{\bm{\mathcal{E}}}=1000,
\lambda_{\bm{\mathcal{P}}}=1000,
\lambda_{\bm{\mathcal{W}}}=1$ and are reduced by half at 15k, 30k, 45k iteration respectively.

We train our model with Adam optimizer for 70k iterations in total, where the three stages of our training take 4k, 46k and 20k iterations respectively.
The learning rate for blendshapes and LBS weight MLP $\MLP_d$, neural texture, anchor Gaussian parameters and neural warping field are set to $10^{-3}$, which is halved at 30k-th and 60k-th iterations respectively. The learning rate and density control hyperparameters for regular Gaussians are the same as proposed by the original paper~\cite{gaussian_splatting}, except that we use a density gradient threshold of $2.5 \times 10^{-4}$ before we start applying VGG loss, and $8 \times 10^{-3}$ afterward. For every $10$k iterations during the training, we also re-project all anchor Gaussians to the image plane of the canonical image plane, and remove the anchor Guassians that are out of the view frustum. This is to prevent unconstrained anchor Gaussians from applying noisy regularization on the texture warping field. 

Following \cite{pointavatar} and \cite{zheng2022imavatar}, we also add a static bone, which does not take any transformation with the FLAME expression and poses.

As our preprocessing pipeline does not track eye movement, for subjects with significant eye movements in the training frames, i.e., subjects 002 and 005, we do not update the opacity and SH of regular Gaussians in the third stage to prevent undesirable view-dependent artifacts. For subjects where the semantic mask fails, i.e., subject 003, the No MLP texture may contain significant noise in the head region. We hence manually define a rough bounding box for this subject to clean the No MLP texture for self-reenactment and cross-reenactment tasks.

The training takes around 2 hours for each subject on an RTX4080 Ti.

\subsection{Evaluation Details}

Following \cite{pointavatar} and \cite{neuralheadavatar}, we also fine-tune the pre-tracked FLAME expression, pose parameters, camera translation and body landmarks during the training to account for inaccuracies in the preprocessing pipeline. We use Adam optimizer with a learning rate of $10^{-4}$ and optimize them from the 30k-th iteration. For test-time tracking optimization, we only use L2 RGB loss. Since we do not have a direct gradient flowing back from the body texture to the FLAME parameters, we also optimize a translation and rotation offset for the body texture mapping.

\section{Additional Results}
\label{sec:additional_results}

\subsection{Videos}

We strongly encourage the readers to watch the videos containing self-reenactment and cross-reenactment results in the supplementary. 

As shown in the videos, existing methods either fail to model the body properly (INSTA~\cite{INSTA}, SplattingAvatar~\cite{SplattingAvatar:CVPR2024}), or fail to learn the details on head and body (PointAvatar~\cite{pointavatar}). While the pure Gaussian Splatting baseline (GS*) could learn the face and body with much better details, it still learns blurry textures and presents severe artifacts when the subject is moving in extreme head rotation. It is most obvious for the self-reenactment and cross-reenactment videos of subject 005 -- many Gaussians modeling the cloth texture are not well-aligned with each other, as a result, they cannot move naturally with the head motion. In comparison, our method can learn extremely sharp textures with robust performance under novel poses and motions.

However, we note that for subjects with extreme body motion and cloth deformation in the training, such as subject 004 and subject 007, our full version might render avatars where the body parts deform slightly unnaturally.  This is because the texture-warping MLP overfits to the noise in the cloth deformation. We note that for those subjects, our No MLP version actually provides a more pleasant and natural rendering.


\subsection{Ablation}

\begin{table}[t]
\small
\centering
\singlespacing
\tabcolsep=0.06cm
\begin{NiceTabular}{l | lll | lll | lll}
\toprule
\multicolumn{1}{c}{}  & \multicolumn{3}{c|}{002}  & \multicolumn{3}{c|}{005} & \multicolumn{3}{c}{007} \\
{} &        \scalebox{0.8}{PSNR} &      \scalebox{0.8}{SSIM} &      \scalebox{0.8}{LPIPS} &        \scalebox{0.8}{PSNR} &       \scalebox{0.8}{SSIM} &      \scalebox{0.8}{LPIPS} &        \scalebox{0.8}{PSNR} &       \scalebox{0.8}{SSIM} &      \scalebox{0.8}{LPIPS}\\
\midrule
No Anchor Loss &           24.96 &          .910 &           .088 &           22.91 &           .854 &           .117 &           19.30 &           .773 &           .134 \\
No Warp Loss   &   \first{32.86} &  \first{.949} &   \first{.041} &  \second{24.19} &  \second{.891} &  \second{.081} &  \second{22.74} &  \second{.848} &  \second{.076} \\
Ours           &  \second{31.98} &  \first{.949} &  \second{.042} &   \first{24.48} &   \first{.895} &   \first{.074} &   \first{23.26} &   \first{.856} &   \first{.074} \\
\bottomrule
\end{NiceTabular}
    \caption{\textbf{Quatitative ablation.} We show the anchor constraint is necessary for learning sharp and correct body texture. While the warp loss might not necessarily improve the performance for the self-reenactment task, it is needed for cross-reenactment with out-of-distribution poses.
    }
    \label{tab:abla}
\end{table}

\begin{figure*}
    \centering
    \includegraphics[width=0.75\textwidth]{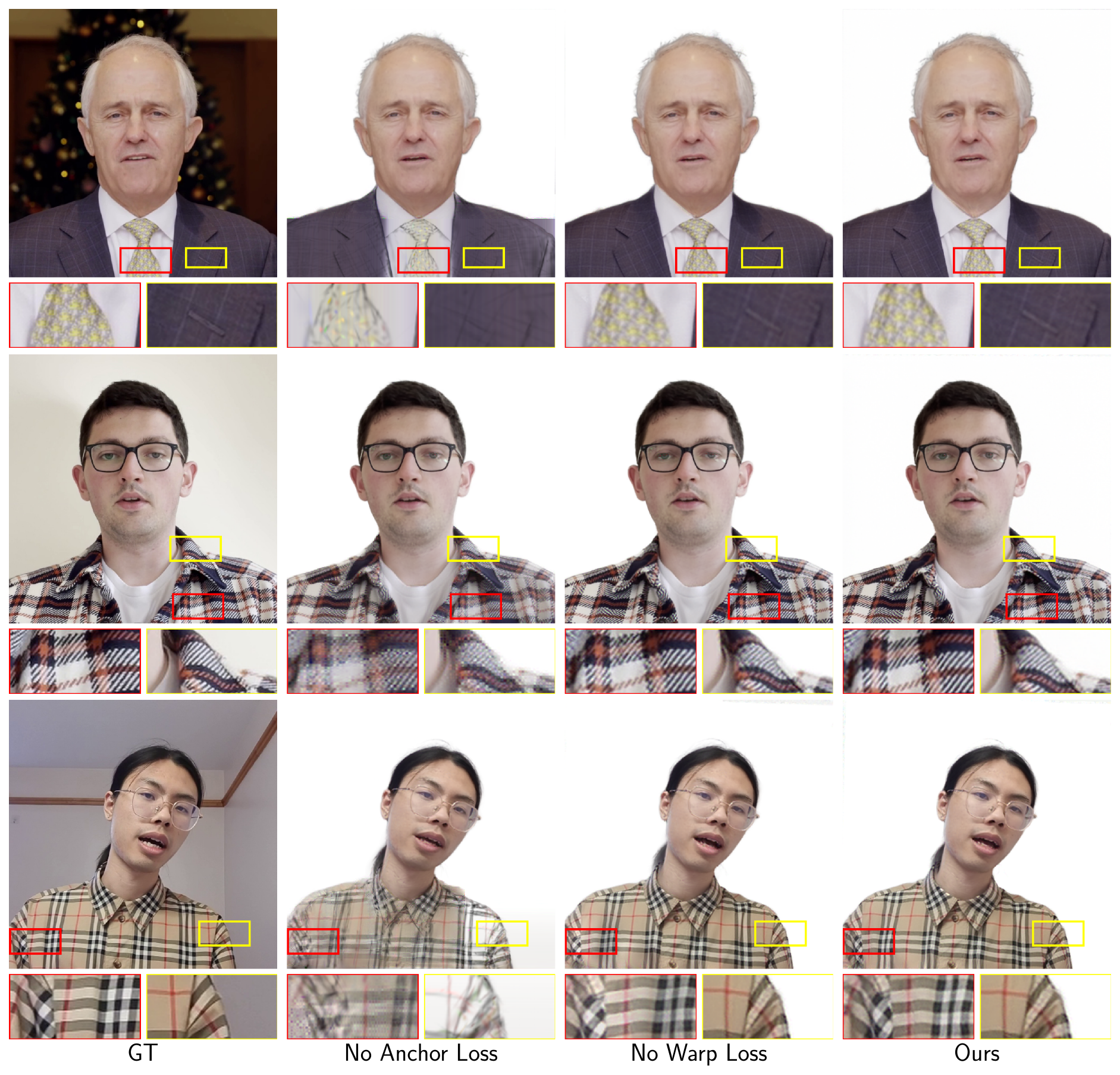}
    \caption{\textbf{Qualitative Ablation.}}
    \label{fig:abla_sup}
\end{figure*}


Additional ablation results are presented in Table~\ref{tab:abla} and Figure~\ref{fig:abla_sup}, demonstrating the critical role of the anchor loss in achieving sharp and precise textures. 
Although the warp loss $\mathcal{L}_{warp}$ does not necessarily improve the numerical metrics for the self-reenactment task, Fig~\ref{fig:abla_reenact} illustrates its importance in preventing the significant failure when combining neural warping with additional Euclidean transformation. 

\subsection{Texture Cleaning}

\begin{figure*}
    \centering
    \includegraphics[width=0.85\textwidth]{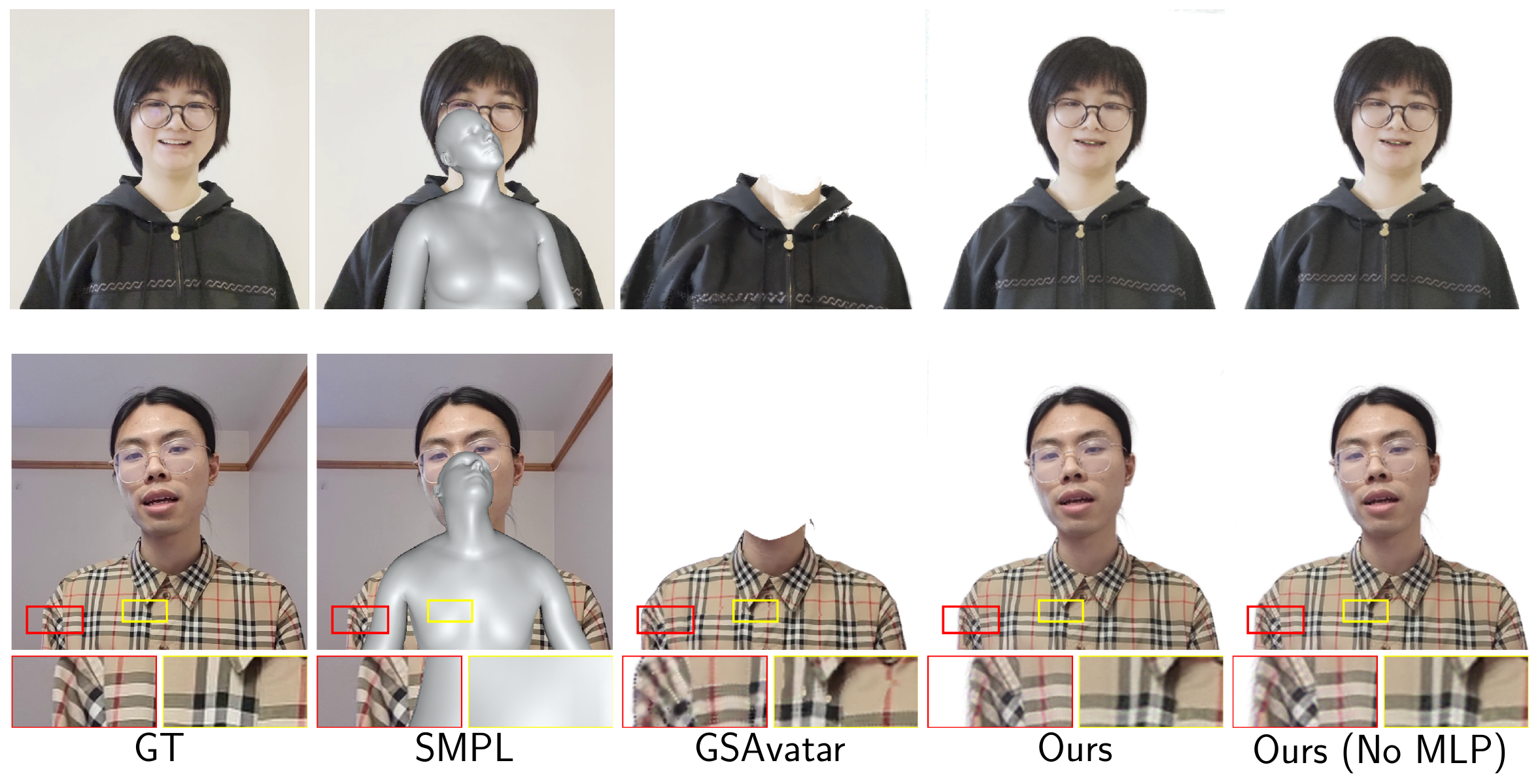}
    \caption{\textbf{Qualitative comparison with full body avatar methods.} Due to the limited landmarks available on the shoulders and chest, existing SMPL tracking methods fail to obtain correct SMPL parameters. Fully body neural avatars that rely on SMPL hence fail to learn accurate and robust body. While our method does not include SMPL 3DMM, the use of static virtual bone and neural texture warping allow us to learn the body texture accurately.}
    \label{fig:body}
\end{figure*}

\begin{figure}
    \centering
    \includegraphics[width=0.45\textwidth]{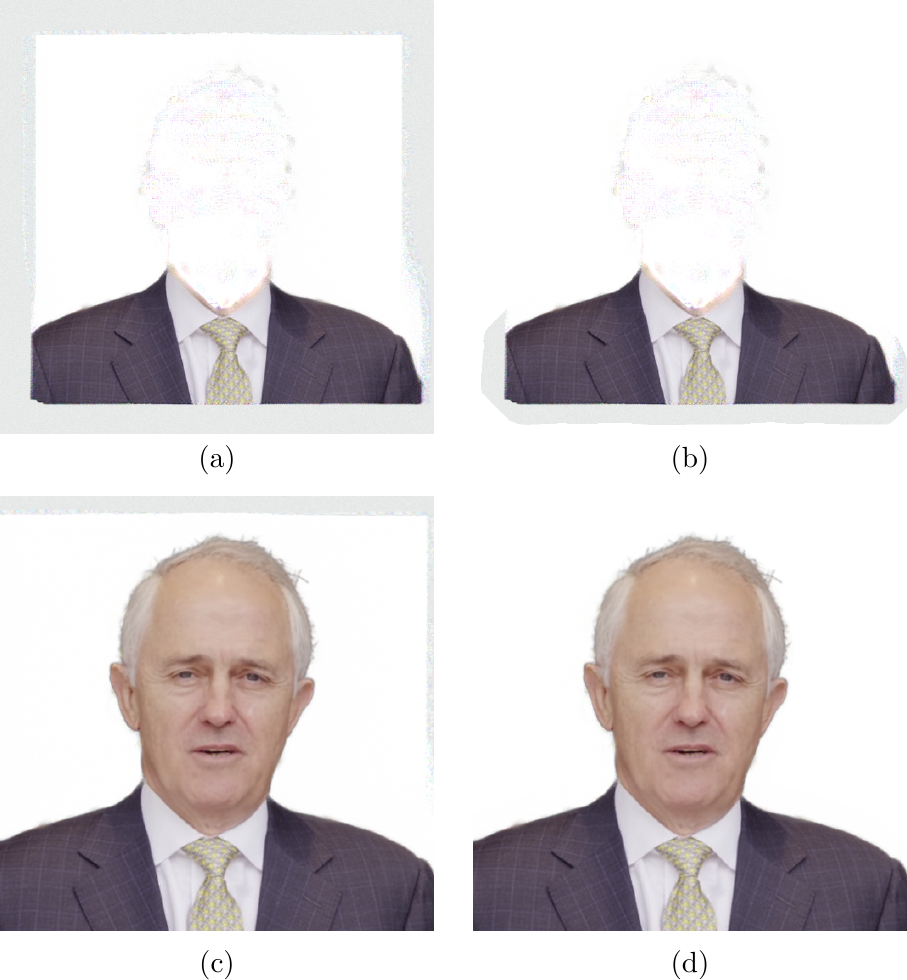}
    \caption{\textbf{Texture cleaning.} We show the body texture without masking (a) and with cleaning (b), as well as the rendering without texture cleaning (c) and with texture cleaning (d).}
    \label{fig:clean_texture}
\end{figure}

When distilling the pose-dependent fine texture into the coarse texture for our no MLP version, we utilized DeepLabV3~\cite{deeplabv3} to obtain a coarse mask of the background and set the values of those pixels to 1. This is needed because the body texture contains a padding region to account for the body part that is moving in and out during the video. A majority section of the padding, especially the padding region on the top the left and right sides, are rarely used and trained during optimization. As a result, the fine texture colors obtained in those regions can produce noisy artifacts; see Fig~\ref{fig:clean_texture}.

\subsection{Comparison with Full Body Avatars}

\begin{table}[t]
\small
\centering
\singlespacing
\tabcolsep=0.06cm
\begin{NiceTabular}{l | lll | lll}
\toprule
\multicolumn{1}{c}{}  & \multicolumn{3}{c|}{004}  & \multicolumn{3}{c}{007} \\
{} &        \scalebox{0.8}{PSNR} &      \scalebox{0.8}{SSIM} &      \scalebox{0.8}{LPIPS} &        \scalebox{0.8}{PSNR} &       \scalebox{0.8}{SSIM} &      \scalebox{0.8}{LPIPS} \\
\midrule
GSAvatar    &           17.08 &           .811 &           .178 &           16.64 &           .744 &           .143 \\
Ours        &   \first{26.82} &   \first{.887} &  \first{.094} &   \first{23.87} &   \first{.885} &   \first{.052} \\
Ours No MLP &  \second{26.70} &  \second{.885} &   \first{.094} &  \second{22.43} &  \second{.861} &  \second{.056} \\
\bottomrule
\end{NiceTabular}
    \caption{\textbf{Body Only Quantitative Comparison with Full Body Avatars.} We show that existing full body neural avatar methods that rely on SMPL deformation perform significantly worse than our methods. Metrics are computed after masking out the background and head regions.
    }
    \label{tab:full_body}
\end{table}

To verify our choice of driving anchor Gaussians only with head 3DMM (FLAME), we select two subjects that show a larger portion of the upper body and compare our method with GSAvatar, a Gaussian Splatting based full body neural avatar methods that deform the representation based on SMPL~\cite{hu2024gaussianavatar_human}. As the code release of GSAvatar only supports SMPL instead of SMPLX, we simply use semantic masks to remove the head region during the training and compare only the reconstruction quality of the body part. As shown in Tab~\ref{tab:full_body} and Fig~\ref{fig:body}, since the existing SMPL tracking methods for monocular videos are developed only for views that include the whole body, the fitted SMPL is significantly misaligned with the GT~\cite{ROMP}, even after fine-tuning during Gaussian optimization. As a result, the clothed body reconstructed by GSAvatar presents several artifacts under novel poses and are significantly misaligned the GT. Our method is able to reconstruct the chest and shoulders with much better quality and accuracy. We would also like to note that, although we do not include body 3DMM in our method, due to the usage of virtual static bone, technically speaking, the effect is exactly the same as have a SMPLX 3DMM where the body and hand parts (SMPLX and MANO) are kept static during the whole sequences.

\subsection{Limitations}


Although we propose a no MLP version that is able to render at novel poses with 130 FPS, as it completely relies on rigid homography transformation to map body texture to the view space, it is unable to model any non-rigid deformation in the body. In addition, for sequences with extreme head rotations, it might move the body in a way that is not exactly aligned with the ground truth, as shown in the supplementary videos. 
However, we observe that the results produced with this no MLP version still present a faithful rendering. 
For cases where the non-rigid body deformation is important, we recommend the use of the full version, whose rendering speed is around 70 FPS and can be further optimized by caching the fine texture only.

\section{Ethics}
\label{sec:ethics}

We captured 4 human subjects with mobile phones for our experiments. All subjects have signed consent forms for using the captured video in this research and publication. We will release the data for subjects with permission.

Our method constructs faithful and animatable head avatars and can be used to generate videos of real people performing synthetic poses and expressions. We do not condone any misuse of our work to generate fake content of any person with the intent of spreading misinformation or tarnishing their reputation.

\bibliographystyle{ACM-Reference-Format}
\bibliography{sample-bibliography}


\begin{thebibliography}{47}


\ifx \showCODEN    \undefined \def \showCODEN     #1{\unskip}     \fi
\ifx \showDOI      \undefined \def \showDOI       #1{#1}\fi
\ifx \showISBNx    \undefined \def \showISBNx     #1{\unskip}     \fi
\ifx \showISBNxiii \undefined \def \showISBNxiii  #1{\unskip}     \fi
\ifx \showISSN     \undefined \def \showISSN      #1{\unskip}     \fi
\ifx \showLCCN     \undefined \def \showLCCN      #1{\unskip}     \fi
\ifx \shownote     \undefined \def \shownote      #1{#1}          \fi
\ifx \showarticletitle \undefined \def \showarticletitle #1{#1}   \fi
\ifx \showURL      \undefined \def \showURL       {\relax}        \fi
\providecommand\bibfield[2]{#2}
\providecommand\bibinfo[2]{#2}
\providecommand\natexlab[1]{#1}
\providecommand\showeprint[2][]{arXiv:#2}

\bibitem[Buehler et~al\mbox{.}(2021)]%
        {varitex}
\bibfield{author}{\bibinfo{person}{Marcel~C. Buehler}, \bibinfo{person}{Abhimitra Meka}, \bibinfo{person}{Gengyan Li}, \bibinfo{person}{Thabo Beeler}, {and} \bibinfo{person}{Otmar Hilliges}.} \bibinfo{year}{2021}\natexlab{}.
\newblock \showarticletitle{VariTex: Variational Neural Face Textures}. In \bibinfo{booktitle}{\emph{Proceedings of the IEEE/CVF International Conference on Computer Vision}}.
\newblock


\bibitem[Casiez et~al\mbox{.}(2012)]%
        {oneeurofilter}
\bibfield{author}{\bibinfo{person}{G\'{e}ry Casiez}, \bibinfo{person}{Nicolas Roussel}, {and} \bibinfo{person}{Daniel Vogel}.} \bibinfo{year}{2012}\natexlab{}.
\newblock \showarticletitle{1 € filter: a simple speed-based low-pass filter for noisy input in interactive systems}. In \bibinfo{booktitle}{\emph{Proceedings of the SIGCHI Conference on Human Factors in Computing Systems}} (Austin, Texas, USA) \emph{(\bibinfo{series}{CHI '12})}. \bibinfo{publisher}{Association for Computing Machinery}, \bibinfo{address}{New York, NY, USA}, \bibinfo{pages}{2527–2530}.
\newblock
\showISBNx{9781450310154}
\urldef\tempurl%
\url{https://doi.org/10.1145/2207676.2208639}
\showDOI{\tempurl}


\bibitem[Chen et~al\mbox{.}(2017)]%
        {deeplabv3}
\bibfield{author}{\bibinfo{person}{Liang-Chieh Chen}, \bibinfo{person}{George Papandreou}, \bibinfo{person}{Florian Schroff}, {and} \bibinfo{person}{Hartwig Adam}.} \bibinfo{year}{2017}\natexlab{}.
\newblock \showarticletitle{Rethinking Atrous Convolution for Semantic Image Segmentation}.
\newblock  (\bibinfo{date}{06} \bibinfo{year}{2017}).
\newblock


\bibitem[Chen et~al\mbox{.}(2023)]%
        {chen2023monogaussianavatar}
\bibfield{author}{\bibinfo{person}{Yufan Chen}, \bibinfo{person}{Lizhen Wang}, \bibinfo{person}{Qijing Li}, \bibinfo{person}{Hongjiang Xiao}, \bibinfo{person}{Shengping Zhang}, \bibinfo{person}{Hongxun Yao}, {and} \bibinfo{person}{Yebin Liu}.} \bibinfo{year}{2023}\natexlab{}.
\newblock \showarticletitle{MonoGaussianAvatar: Monocular Gaussian Point-based Head Avatar}.
\newblock \bibinfo{journal}{\emph{arXiv}} (\bibinfo{year}{2023}).
\newblock


\bibitem[Dhamo et~al\mbox{.}(2023)]%
        {dhamo2023headgas}
\bibfield{author}{\bibinfo{person}{Helisa Dhamo}, \bibinfo{person}{Yinyu Nie}, \bibinfo{person}{Arthur Moreau}, \bibinfo{person}{Jifei Song}, \bibinfo{person}{Richard Shaw}, \bibinfo{person}{Yiren Zhou}, {and} \bibinfo{person}{Eduardo Pérez-Pellitero}.} \bibinfo{year}{2023}\natexlab{}.
\newblock \bibinfo{title}{HeadGaS: Real-Time Animatable Head Avatars via 3D Gaussian Splatting}.
\newblock
\newblock
\showeprint[arxiv]{2312.02902}~[cs.CV]


\bibitem[Feng et~al\mbox{.}(2021)]%
        {DECA:Siggraph2021}
\bibfield{author}{\bibinfo{person}{Yao Feng}, \bibinfo{person}{Haiwen Feng}, \bibinfo{person}{Michael~J. Black}, {and} \bibinfo{person}{Timo Bolkart}.} \bibinfo{year}{2021}\natexlab{}.
\newblock \showarticletitle{Learning an Animatable Detailed {3D} Face Model from In-The-Wild Images}.
\newblock \bibinfo{journal}{\emph{ACM Transactions on Graphics, (Proc. SIGGRAPH)}} \bibinfo{volume}{40}, \bibinfo{number}{8}.
\newblock
\urldef\tempurl%
\url{https://doi.org/10.1145/3450626.3459936}
\showURL{%
\tempurl}


\bibitem[Fischler and Bolles(1981)]%
        {ransac}
\bibfield{author}{\bibinfo{person}{Martin~A. Fischler} {and} \bibinfo{person}{Robert~C. Bolles}.} \bibinfo{year}{1981}\natexlab{}.
\newblock \showarticletitle{Random sample consensus: a paradigm for model fitting with applications to image analysis and automated cartography}.
\newblock \bibinfo{journal}{\emph{Commun. ACM}} \bibinfo{volume}{24}, \bibinfo{number}{6} (\bibinfo{date}{jun} \bibinfo{year}{1981}), \bibinfo{pages}{381–395}.
\newblock
\showISSN{0001-0782}
\urldef\tempurl%
\url{https://doi.org/10.1145/358669.358692}
\showDOI{\tempurl}


\bibitem[Gafni et~al\mbox{.}(2021)]%
        {dynerf_for_face}
\bibfield{author}{\bibinfo{person}{Guy Gafni}, \bibinfo{person}{Justus Thies}, \bibinfo{person}{Michael Zollh{\"o}fer}, {and} \bibinfo{person}{Matthias Nie{\ss}ner}.} \bibinfo{year}{2021}\natexlab{}.
\newblock \showarticletitle{Dynamic Neural Radiance Fields for Monocular 4D Facial Avatar Reconstruction}. In \bibinfo{booktitle}{\emph{Proceedings of the IEEE/CVF Conference on Computer Vision and Pattern Recognition (CVPR)}}. \bibinfo{pages}{8649--8658}.
\newblock


\bibitem[Gao et~al\mbox{.}(2022)]%
        {Gao2022nerfblendshape}
\bibfield{author}{\bibinfo{person}{Xuan Gao}, \bibinfo{person}{Chenglai Zhong}, \bibinfo{person}{Jun Xiang}, \bibinfo{person}{Yang Hong}, \bibinfo{person}{Yudong Guo}, {and} \bibinfo{person}{Juyong Zhang}.} \bibinfo{year}{2022}\natexlab{}.
\newblock \showarticletitle{Reconstructing Personalized Semantic Facial NeRF Models From Monocular Video}.
\newblock \bibinfo{journal}{\emph{ACM Transactions on Graphics (Proceedings of SIGGRAPH Asia)}} \bibinfo{volume}{41}, \bibinfo{number}{6} (\bibinfo{year}{2022}).
\newblock
\urldef\tempurl%
\url{https://doi.org/10.1145/3550454.3555501}
\showDOI{\tempurl}


\bibitem[Gerig et~al\mbox{.}(2017)]%
        {bfm}
\bibfield{author}{\bibinfo{person}{Thomas Gerig}, \bibinfo{person}{Andreas Forster}, \bibinfo{person}{Clemens Blumer}, \bibinfo{person}{Bernhard Egger}, \bibinfo{person}{Marcel L{\"{u}}thi}, \bibinfo{person}{Sandro Sch{\"{o}}nborn}, {and} \bibinfo{person}{Thomas Vetter}.} \bibinfo{year}{2017}\natexlab{}.
\newblock \showarticletitle{Morphable Face Models - An Open Framework}.
\newblock \bibinfo{journal}{\emph{CoRR}}  \bibinfo{volume}{abs/1709.08398} (\bibinfo{year}{2017}).
\newblock
\showeprint[arXiv]{1709.08398}
\urldef\tempurl%
\url{http://arxiv.org/abs/1709.08398}
\showURL{%
\tempurl}


\bibitem[Grassal et~al\mbox{.}(2021)]%
        {neuralheadavatar}
\bibfield{author}{\bibinfo{person}{Philip-William Grassal}, \bibinfo{person}{Malte Prinzler}, \bibinfo{person}{Titus Leistner}, \bibinfo{person}{Carsten Rother}, \bibinfo{person}{Matthias Nie{\ss}ner}, {and} \bibinfo{person}{Justus Thies}.} \bibinfo{year}{2021}\natexlab{}.
\newblock \showarticletitle{Neural Head Avatars from Monocular RGB Videos}.
\newblock \bibinfo{journal}{\emph{arXiv preprint arXiv:2112.01554}} (\bibinfo{year}{2021}).
\newblock


\bibitem[Gropp et~al\mbox{.}(2020)]%
        {icml2020_2086}
\bibfield{author}{\bibinfo{person}{Amos Gropp}, \bibinfo{person}{Lior Yariv}, \bibinfo{person}{Niv Haim}, \bibinfo{person}{Matan Atzmon}, {and} \bibinfo{person}{Yaron Lipman}.} \bibinfo{year}{2020}\natexlab{}.
\newblock \showarticletitle{Implicit Geometric Regularization for Learning Shapes}.
\newblock In \bibinfo{booktitle}{\emph{Proceedings of Machine Learning and Systems 2020}}. \bibinfo{pages}{3569--3579}.
\newblock


\bibitem[Hu et~al\mbox{.}(2024)]%
        {hu2024gaussianavatar_human}
\bibfield{author}{\bibinfo{person}{Liangxiao Hu}, \bibinfo{person}{Hongwen Zhang}, \bibinfo{person}{Yuxiang Zhang}, \bibinfo{person}{Boyao Zhou}, \bibinfo{person}{Boning Liu}, \bibinfo{person}{Shengping Zhang}, {and} \bibinfo{person}{Liqiang Nie}.} \bibinfo{year}{2024}\natexlab{}.
\newblock \showarticletitle{GaussianAvatar: Towards Realistic Human Avatar Modeling from a Single Video via Animatable 3D Gaussians}. In \bibinfo{booktitle}{\emph{IEEE/CVF Conference on Computer Vision and Pattern Recognition (CVPR)}}.
\newblock


\bibitem[Hu and Liu(2023)]%
        {hu2023gauhuman}
\bibfield{author}{\bibinfo{person}{Shoukang Hu} {and} \bibinfo{person}{Ziwei Liu}.} \bibinfo{year}{2023}\natexlab{}.
\newblock \showarticletitle{GauHuman: Articulated Gaussian Splatting from Monocular Human Videos}.
\newblock \bibinfo{journal}{\emph{arXiv preprint arXiv:}} (\bibinfo{year}{2023}).
\newblock


\bibitem[Jiang et~al\mbox{.}(2022)]%
        {jiang2022neuman}
\bibfield{author}{\bibinfo{person}{Wei Jiang}, \bibinfo{person}{Kwang~Moo Yi}, \bibinfo{person}{Golnoosh Samei}, \bibinfo{person}{Oncel Tuzel}, {and} \bibinfo{person}{Anurag Ranjan}.} \bibinfo{year}{2022}\natexlab{}.
\newblock \showarticletitle{NeuMan: Neural Human Radiance Field from a Single Video}. In \bibinfo{booktitle}{\emph{Proceedings of the European conference on computer vision (ECCV)}}.
\newblock


\bibitem[Johnson et~al\mbox{.}(2016)]%
        {vgg_loss}
\bibfield{author}{\bibinfo{person}{Justin Johnson}, \bibinfo{person}{Alexandre Alahi}, {and} \bibinfo{person}{Li Fei-Fei}.} \bibinfo{year}{2016}\natexlab{}.
\newblock \showarticletitle{Perceptual Losses for Real-Time Style Transfer and Super-Resolution}. In \bibinfo{booktitle}{\emph{Computer Vision -- ECCV 2016}}, \bibfield{editor}{\bibinfo{person}{Bastian Leibe}, \bibinfo{person}{Jiri Matas}, \bibinfo{person}{Nicu Sebe}, {and} \bibinfo{person}{Max Welling}} (Eds.). \bibinfo{publisher}{Springer International Publishing}, \bibinfo{address}{Cham}, \bibinfo{pages}{694--711}.
\newblock
\showISBNx{978-3-319-46475-6}


\bibitem[Kerbl et~al\mbox{.}(2023)]%
        {gaussian_splatting}
\bibfield{author}{\bibinfo{person}{Bernhard Kerbl}, \bibinfo{person}{Georgios Kopanas}, \bibinfo{person}{Thomas Leimk{\"u}hler}, {and} \bibinfo{person}{George Drettakis}.} \bibinfo{year}{2023}\natexlab{}.
\newblock \showarticletitle{3D Gaussian Splatting for Real-Time Radiance Field Rendering}.
\newblock \bibinfo{journal}{\emph{ACM Transactions on Graphics}} \bibinfo{volume}{42}, \bibinfo{number}{4} (\bibinfo{date}{July} \bibinfo{year}{2023}).
\newblock
\urldef\tempurl%
\url{https://repo-sam.inria.fr/fungraph/3d-gaussian-splatting/}
\showURL{%
\tempurl}


\bibitem[Khakhulin et~al\mbox{.}(2022)]%
        {Khakhulin2022ROME}
\bibfield{author}{\bibinfo{person}{Taras Khakhulin}, \bibinfo{person}{Vanessa Sklyarova}, \bibinfo{person}{Victor Lempitsky}, {and} \bibinfo{person}{Egor Zakharov}.} \bibinfo{year}{2022}\natexlab{}.
\newblock \showarticletitle{Realistic One-shot Mesh-based Head Avatars}. In \bibinfo{booktitle}{\emph{European Conference of Computer vision (ECCV)}}.
\newblock


\bibitem[Kim et~al\mbox{.}(2018)]%
        {kim2018deep_video_portrait}
\bibfield{author}{\bibinfo{person}{Hyeongwoo Kim}, \bibinfo{person}{Pablo Garrido}, \bibinfo{person}{Ayush Tewari}, \bibinfo{person}{Weipeng Xu}, \bibinfo{person}{Justus Thies}, \bibinfo{person}{Matthias Nie{\ss}ner}, \bibinfo{person}{Patrick P{\'e}rez}, \bibinfo{person}{Christian Richardt}, \bibinfo{person}{Michael Zoll{\"o}fer}, {and} \bibinfo{person}{Christian Theobalt}.} \bibinfo{year}{2018}\natexlab{}.
\newblock \showarticletitle{Deep Video Portraits}.
\newblock \bibinfo{journal}{\emph{ACM Transactions on Graphics (TOG)}} \bibinfo{volume}{37}, \bibinfo{number}{4} (\bibinfo{year}{2018}), \bibinfo{pages}{163}.
\newblock


\bibitem[Kirschstein et~al\mbox{.}(2023)]%
        {kirschstein2023nersemble}
\bibfield{author}{\bibinfo{person}{Tobias Kirschstein}, \bibinfo{person}{Shenhan Qian}, \bibinfo{person}{Simon Giebenhain}, \bibinfo{person}{Tim Walter}, {and} \bibinfo{person}{Matthias Nie\ss{}ner}.} \bibinfo{year}{2023}\natexlab{}.
\newblock \showarticletitle{NeRSemble: Multi-View Radiance Field Reconstruction of Human Heads}.
\newblock \bibinfo{journal}{\emph{ACM Trans. Graph.}} \bibinfo{volume}{42}, \bibinfo{number}{4}, Article \bibinfo{articleno}{161} (\bibinfo{date}{jul} \bibinfo{year}{2023}), \bibinfo{numpages}{14}~pages.
\newblock
\showISSN{0730-0301}
\urldef\tempurl%
\url{https://doi.org/10.1145/3592455}
\showDOI{\tempurl}


\bibitem[Kocabas et~al\mbox{.}(2023)]%
        {hugs}
\bibfield{author}{\bibinfo{person}{Muhammed Kocabas}, \bibinfo{person}{Rick Chang}, \bibinfo{person}{James Gabriel}, \bibinfo{person}{Oncel Tuzel}, {and} \bibinfo{person}{Anurag Ranjan}.} \bibinfo{year}{2023}\natexlab{}.
\newblock \bibinfo{title}{HUGS: Human Gaussian Splats}.
\newblock
\newblock
\urldef\tempurl%
\url{https://arxiv.org/abs/2311.17910}
\showURL{%
\tempurl}


\bibitem[Koujan et~al\mbox{.}(2020)]%
        {head2head2020}
\bibfield{author}{\bibinfo{person}{M. Koujan}, \bibinfo{person}{M. Doukas}, \bibinfo{person}{A. Roussos}, {and} \bibinfo{person}{S. Zafeiriou}.} \bibinfo{year}{2020}\natexlab{}.
\newblock \showarticletitle{Head2Head: Video-Based Neural Head Synthesis}. In \bibinfo{booktitle}{\emph{2020 15th IEEE International Conference on Automatic Face and Gesture Recognition (FG 2020) (FG)}}. \bibinfo{publisher}{IEEE Computer Society}, \bibinfo{address}{Los Alamitos, CA, USA}, \bibinfo{pages}{319--326}.
\newblock
\urldef\tempurl%
\url{https://doi.org/10.1109/FG47880.2020.00048}
\showDOI{\tempurl}


\bibitem[Lei et~al\mbox{.}(2023)]%
        {lei2023gart}
\bibfield{author}{\bibinfo{person}{Jiahui Lei}, \bibinfo{person}{Yufu Wang}, \bibinfo{person}{Georgios Pavlakos}, \bibinfo{person}{Lingjie Liu}, {and} \bibinfo{person}{Kostas Daniilidis}.} \bibinfo{year}{2023}\natexlab{}.
\newblock \bibinfo{title}{GART: Gaussian Articulated Template Models}.
\newblock
\newblock
\showeprint[arxiv]{2311.16099}~[cs.CV]


\bibitem[Li et~al\mbox{.}(2017)]%
        {FLAME:SiggraphAsia2017}
\bibfield{author}{\bibinfo{person}{Tianye Li}, \bibinfo{person}{Timo Bolkart}, \bibinfo{person}{Michael.~J. Black}, \bibinfo{person}{Hao Li}, {and} \bibinfo{person}{Javier Romero}.} \bibinfo{year}{2017}\natexlab{}.
\newblock \showarticletitle{Learning a model of facial shape and expression from {4D} scans}.
\newblock \bibinfo{journal}{\emph{ACM Transactions on Graphics, (Proc. SIGGRAPH Asia)}} \bibinfo{volume}{36}, \bibinfo{number}{6} (\bibinfo{year}{2017}), \bibinfo{pages}{194:1--194:17}.
\newblock
\urldef\tempurl%
\url{https://doi.org/10.1145/3130800.3130813}
\showURL{%
\tempurl}


\bibitem[Li et~al\mbox{.}(2024)]%
        {li2024animatablegaussians_human}
\bibfield{author}{\bibinfo{person}{Zhe Li}, \bibinfo{person}{Zerong Zheng}, \bibinfo{person}{Lizhen Wang}, {and} \bibinfo{person}{Yebin Liu}.} \bibinfo{year}{2024}\natexlab{}.
\newblock \showarticletitle{Animatable Gaussians: Learning Pose-dependent Gaussian Maps for High-fidelity Human Avatar Modeling}. In \bibinfo{booktitle}{\emph{Proceedings of the IEEE/CVF Conference on Computer Vision and Pattern Recognition (CVPR)}}.
\newblock


\bibitem[Liu et~al\mbox{.}(2024)]%
        {liu24-GVA}
\bibfield{author}{\bibinfo{person}{Xinqi Liu}, \bibinfo{person}{Chenming Wu}, \bibinfo{person}{Jialun Liu}, \bibinfo{person}{Xing Liu}, \bibinfo{person}{Chen Zhao}, \bibinfo{person}{Haocheng Feng}, \bibinfo{person}{Errui Ding}, {and} \bibinfo{person}{Jingdong Wang}.} \bibinfo{year}{2024}\natexlab{}.
\newblock \showarticletitle{GVA: Reconstructing Vivid 3D Gaussian Avatars from Monocular Videos}.
\newblock \bibinfo{journal}{\emph{Arxiv}} (\bibinfo{year}{2024}).
\newblock


\bibitem[Loper et~al\mbox{.}(2015)]%
        {SMPL:2015}
\bibfield{author}{\bibinfo{person}{Matthew Loper}, \bibinfo{person}{Naureen Mahmood}, \bibinfo{person}{Javier Romero}, \bibinfo{person}{Gerard Pons-Moll}, {and} \bibinfo{person}{Michael~J. Black}.} \bibinfo{year}{2015}\natexlab{}.
\newblock \showarticletitle{{SMPL}: A Skinned Multi-Person Linear Model}.
\newblock \bibinfo{journal}{\emph{ACM Trans. Graphics (Proc. SIGGRAPH Asia)}} \bibinfo{volume}{34}, \bibinfo{number}{6} (\bibinfo{date}{Oct.} \bibinfo{year}{2015}), \bibinfo{pages}{248:1--248:16}.
\newblock


\bibitem[Mescheder et~al\mbox{.}(2019)]%
        {occupancy_network}
\bibfield{author}{\bibinfo{person}{L. Mescheder}, \bibinfo{person}{M. Oechsle}, \bibinfo{person}{M. Niemeyer}, \bibinfo{person}{S. Nowozin}, {and} \bibinfo{person}{A. Geiger}.} \bibinfo{year}{2019}\natexlab{}.
\newblock \showarticletitle{Occupancy Networks: Learning 3D Reconstruction in Function Space}. In \bibinfo{booktitle}{\emph{2019 IEEE/CVF Conference on Computer Vision and Pattern Recognition (CVPR)}}. \bibinfo{publisher}{IEEE Computer Society}, \bibinfo{address}{Los Alamitos, CA, USA}, \bibinfo{pages}{4455--4465}.
\newblock
\urldef\tempurl%
\url{https://doi.org/10.1109/CVPR.2019.00459}
\showDOI{\tempurl}


\bibitem[Mildenhall et~al\mbox{.}(2020)]%
        {nerf}
\bibfield{author}{\bibinfo{person}{Ben Mildenhall}, \bibinfo{person}{Pratul~P. Srinivasan}, \bibinfo{person}{Matthew Tancik}, \bibinfo{person}{Jonathan~T. Barron}, \bibinfo{person}{Ravi Ramamoorthi}, {and} \bibinfo{person}{Ren Ng}.} \bibinfo{year}{2020}\natexlab{}.
\newblock \showarticletitle{NeRF: Representing Scenes as Neural Radiance Fields for View Synthesis}. In \bibinfo{booktitle}{\emph{ECCV}}.
\newblock


\bibitem[M\"uller et~al\mbox{.}(2022)]%
        {instantNGP}
\bibfield{author}{\bibinfo{person}{Thomas M\"uller}, \bibinfo{person}{Alex Evans}, \bibinfo{person}{Christoph Schied}, {and} \bibinfo{person}{Alexander Keller}.} \bibinfo{year}{2022}\natexlab{}.
\newblock \showarticletitle{Instant Neural Graphics Primitives with a Multiresolution Hash Encoding}.
\newblock \bibinfo{journal}{\emph{ACM Trans. Graph.}} \bibinfo{volume}{41}, \bibinfo{number}{4}, Article \bibinfo{articleno}{102} (\bibinfo{date}{July} \bibinfo{year}{2022}), \bibinfo{numpages}{15}~pages.
\newblock
\urldef\tempurl%
\url{https://doi.org/10.1145/3528223.3530127}
\showDOI{\tempurl}


\bibitem[Park et~al\mbox{.}(2021)]%
        {park2021hypernerf}
\bibfield{author}{\bibinfo{person}{Keunhong Park}, \bibinfo{person}{Utkarsh Sinha}, \bibinfo{person}{Peter Hedman}, \bibinfo{person}{Jonathan~T. Barron}, \bibinfo{person}{Sofien Bouaziz}, \bibinfo{person}{Dan~B Goldman}, \bibinfo{person}{Ricardo Martin-Brualla}, {and} \bibinfo{person}{Steven~M. Seitz}.} \bibinfo{year}{2021}\natexlab{}.
\newblock \showarticletitle{HyperNeRF: A Higher-Dimensional Representation for Topologically Varying Neural Radiance Fields}.
\newblock \bibinfo{journal}{\emph{ACM Trans. Graph.}} \bibinfo{volume}{40}, \bibinfo{number}{6}, Article \bibinfo{articleno}{238} (\bibinfo{date}{dec} \bibinfo{year}{2021}).
\newblock


\bibitem[Qi et~al\mbox{.}(2017)]%
        {qi2017pointnetplusplus}
\bibfield{author}{\bibinfo{person}{Charles~R Qi}, \bibinfo{person}{Li Yi}, \bibinfo{person}{Hao Su}, {and} \bibinfo{person}{Leonidas~J Guibas}.} \bibinfo{year}{2017}\natexlab{}.
\newblock \showarticletitle{PointNet++: Deep Hierarchical Feature Learning on Point Sets in a Metric Space}.
\newblock \bibinfo{journal}{\emph{arXiv preprint arXiv:1706.02413}} (\bibinfo{year}{2017}).
\newblock


\bibitem[Saito et~al\mbox{.}(2024)]%
        {saito2024rgca}
\bibfield{author}{\bibinfo{person}{Shunsuke Saito}, \bibinfo{person}{Gabriel Schwartz}, \bibinfo{person}{Tomas Simon}, \bibinfo{person}{Junxuan Li}, {and} \bibinfo{person}{Giljoo Nam}.} \bibinfo{year}{2024}\natexlab{}.
\newblock \showarticletitle{Relightable Gaussian Codec Avatars}. In \bibinfo{booktitle}{\emph{CVPR}}.
\newblock


\bibitem[Shao et~al\mbox{.}(2024)]%
        {SplattingAvatar:CVPR2024}
\bibfield{author}{\bibinfo{person}{Zhijing Shao}, \bibinfo{person}{Zhaolong Wang}, \bibinfo{person}{Zhuang Li}, \bibinfo{person}{Duotun Wang}, \bibinfo{person}{Xiangru Lin}, \bibinfo{person}{Yu Zhang}, \bibinfo{person}{Mingming Fan}, {and} \bibinfo{person}{Zeyu Wang}.} \bibinfo{year}{2024}\natexlab{}.
\newblock \showarticletitle{{SplattingAvatar: Realistic Real-Time Human Avatars with Mesh-Embedded Gaussian Splatting}}. In \bibinfo{booktitle}{\emph{Computer Vision and Pattern Recognition (CVPR)}}.
\newblock


\bibitem[Simonyan and Zisserman(2015)]%
        {vgg}
\bibfield{author}{\bibinfo{person}{Karen Simonyan} {and} \bibinfo{person}{Andrew Zisserman}.} \bibinfo{year}{2015}\natexlab{}.
\newblock \showarticletitle{Very Deep Convolutional Networks for Large-Scale Image Recognition}. In \bibinfo{booktitle}{\emph{3rd International Conference on Learning Representations, {ICLR} 2015, San Diego, CA, USA, May 7-9, 2015, Conference Track Proceedings}}, \bibfield{editor}{\bibinfo{person}{Yoshua Bengio} {and} \bibinfo{person}{Yann LeCun}} (Eds.).
\newblock
\urldef\tempurl%
\url{http://arxiv.org/abs/1409.1556}
\showURL{%
\tempurl}


\bibitem[Sun et~al\mbox{.}(2021)]%
        {ROMP}
\bibfield{author}{\bibinfo{person}{Yu Sun}, \bibinfo{person}{Qian Bao}, \bibinfo{person}{Wu Liu}, \bibinfo{person}{Yili Fu}, \bibinfo{person}{Black Michael~J.}, {and} \bibinfo{person}{Tao Mei}.} \bibinfo{year}{2021}\natexlab{}.
\newblock \showarticletitle{{Monocular, One-stage, Regression of Multiple 3D People}}. In \bibinfo{booktitle}{\emph{ICCV}}.
\newblock


\bibitem[Svitov et~al\mbox{.}(2024)]%
        {svitov2024haha}
\bibfield{author}{\bibinfo{person}{David Svitov}, \bibinfo{person}{Pietro Morerio}, \bibinfo{person}{Lourdes Agapito}, {and} \bibinfo{person}{Alessio~Del Bue}.} \bibinfo{year}{2024}\natexlab{}.
\newblock \bibinfo{title}{HAHA: Highly Articulated Gaussian Human Avatars with Textured Mesh Prior}.
\newblock
\newblock
\showeprint[arxiv]{2404.01053}~[cs.CV]


\bibitem[Wang et~al\mbox{.}(2024)]%
        {gaussianblendshape}
\bibfield{author}{\bibinfo{person}{Jie Wang}, \bibinfo{person}{Jiu-Cheng Xie}, \bibinfo{person}{Xianyan Li}, \bibinfo{person}{Feng Xu}, \bibinfo{person}{Chi-Man Pun}, {and} \bibinfo{person}{Hao Gao}.} \bibinfo{year}{2024}\natexlab{}.
\newblock \bibinfo{title}{GaussianHead: High-fidelity Head Avatars with Learnable Gaussian Derivation}.
\newblock
\newblock
\showeprint[arxiv]{2312.01632}~[cs.CV]


\bibitem[Wang et~al\mbox{.}(2021)]%
        {composition_dynerf_for_head}
\bibfield{author}{\bibinfo{person}{Ziyan Wang}, \bibinfo{person}{Timur Bagautdinov}, \bibinfo{person}{Stephen Lombardi}, \bibinfo{person}{Tomas Simon}, \bibinfo{person}{Jason Saragih}, \bibinfo{person}{Jessica Hodgins}, {and} \bibinfo{person}{Michael Zollh¨ofer}.} \bibinfo{year}{2021}\natexlab{}.
\newblock \showarticletitle{Learning Compositional Radiance Fields of Dynamic Human Heads}. In \bibinfo{booktitle}{\emph{Proceedings of the IEEE/CVF Conference on Computer Vision and Pattern Recognition}}. \bibinfo{pages}{5704--5713}.
\newblock


\bibitem[Xiang et~al\mbox{.}(2024)]%
        {xiang2024flashavatar}
\bibfield{author}{\bibinfo{person}{Jun Xiang}, \bibinfo{person}{Xuan Gao}, \bibinfo{person}{Yudong Guo}, {and} \bibinfo{person}{Juyong Zhang}.} \bibinfo{year}{2024}\natexlab{}.
\newblock \showarticletitle{FlashAvatar: High-fidelity Head Avatar with Efficient Gaussian Embedding}. In \bibinfo{booktitle}{\emph{The IEEE Conference on Computer Vision and Pattern Recognition (CVPR)}}.
\newblock


\bibitem[Xu et~al\mbox{.}(2023)]%
        {xu2023avatarmav}
\bibfield{author}{\bibinfo{person}{Yuelang Xu}, \bibinfo{person}{Lizhen Wang}, \bibinfo{person}{Xiaochen Zhao}, \bibinfo{person}{Hongwen Zhang}, {and} \bibinfo{person}{Yebin Liu}.} \bibinfo{year}{2023}\natexlab{}.
\newblock \showarticletitle{AvatarMAV: Fast 3D Head Avatar Reconstruction Using Motion-Aware Neural Voxels}. In \bibinfo{booktitle}{\emph{ACM SIGGRAPH 2023 Conference Proceedings}}.
\newblock


\bibitem[Yang et~al\mbox{.}(2023)]%
        {dwpose}
\bibfield{author}{\bibinfo{person}{Zhendong Yang}, \bibinfo{person}{Ailing Zeng}, \bibinfo{person}{Chun Yuan}, {and} \bibinfo{person}{Yu Li}.} \bibinfo{year}{2023}\natexlab{}.
\newblock \showarticletitle{Effective whole-body pose estimation with two-stages distillation}. In \bibinfo{booktitle}{\emph{Proceedings of the IEEE/CVF International Conference on Computer Vision}}. \bibinfo{pages}{4210--4220}.
\newblock


\bibitem[Zhao et~al\mbox{.}(2024)]%
        {psavatar}
\bibfield{author}{\bibinfo{person}{Zhongyuan Zhao}, \bibinfo{person}{Zhenyu Bao}, \bibinfo{person}{Qing Li}, \bibinfo{person}{Guoping Qiu}, {and} \bibinfo{person}{Kanglin Liu}.} \bibinfo{year}{2024}\natexlab{}.
\newblock \bibinfo{title}{PSAvatar: A Point-based Morphable Shape Model for Real-Time Head Avatar Animation with 3D Gaussian Splatting}.
\newblock
\newblock
\showeprint[arxiv]{2401.12900}~[cs.GR]


\bibitem[Zheng et~al\mbox{.}(2022)]%
        {zheng2022imavatar}
\bibfield{author}{\bibinfo{person}{Yufeng Zheng}, \bibinfo{person}{Victoria~Fernández Abrevaya}, \bibinfo{person}{Marcel~C. Bühler}, \bibinfo{person}{Xu Chen}, \bibinfo{person}{Michael~J. Black}, {and} \bibinfo{person}{Otmar Hilliges}.} \bibinfo{year}{2022}\natexlab{}.
\newblock \showarticletitle{{I} {M} {Avatar}: Implicit Morphable Head Avatars from Videos}. In \bibinfo{booktitle}{\emph{Computer Vision and Pattern Recognition (CVPR)}}.
\newblock


\bibitem[Zheng et~al\mbox{.}(2023)]%
        {pointavatar}
\bibfield{author}{\bibinfo{person}{Yufeng Zheng}, \bibinfo{person}{Wang Yifan}, \bibinfo{person}{Gordon Wetzstein}, \bibinfo{person}{Michael~J. Black}, {and} \bibinfo{person}{Otmar Hilliges}.} \bibinfo{year}{2023}\natexlab{}.
\newblock \showarticletitle{PointAvatar: Deformable Point-based Head Avatars from Videos}. In \bibinfo{booktitle}{\emph{Proceedings of the IEEE/CVF Conference on Computer Vision and Pattern Recognition (CVPR)}}.
\newblock


\bibitem[Zielonka et~al\mbox{.}(2022)]%
        {INSTA}
\bibfield{author}{\bibinfo{person}{Wojciech Zielonka}, \bibinfo{person}{Timo Bolkart}, {and} \bibinfo{person}{Justus Thies}.} \bibinfo{year}{2022}\natexlab{}.
\newblock \showarticletitle{Instant Volumetric Head Avatars}.
\newblock \bibinfo{journal}{\emph{2023 IEEE/CVF Conference on Computer Vision and Pattern Recognition (CVPR)}} (\bibinfo{year}{2022}), \bibinfo{pages}{4574--4584}.
\newblock
\urldef\tempurl%
\url{https://api.semanticscholar.org/CorpusID:253761096}
\showURL{%
\tempurl}


\bibitem[Zwicker et~al\mbox{.}(2001)]%
        {ewa_volume_splatting}
\bibfield{author}{\bibinfo{person}{M. Zwicker}, \bibinfo{person}{H. Pfister}, \bibinfo{person}{J. van Baar}, {and} \bibinfo{person}{M. Gross}.} \bibinfo{year}{2001}\natexlab{}.
\newblock \showarticletitle{EWA volume splatting}. In \bibinfo{booktitle}{\emph{Proceedings Visualization, 2001. VIS '01.}} \bibinfo{pages}{29--538}.
\newblock
\urldef\tempurl%
\url{https://doi.org/10.1109/VISUAL.2001.964490}
\showDOI{\tempurl}


\end{thebibliography}

\end{document}